\documentclass[11pt,a4paper]{article}
\usepackage{times,latexsym}
\usepackage{url}
\usepackage{verbatim}
\usepackage{graphicx}
\usepackage{tabularx}
\usepackage{booktabs}
\usepackage{pifont}
\usepackage{multirow}
\usepackage{ulem}
\usepackage{amsmath,amsfonts}
\usepackage{algorithmic}
\usepackage{algorithm}
\usepackage{array}
\usepackage{mdwlist}
\usepackage[T1]{fontenc}
\usepackage{tcolorbox}
\usepackage{CJK}

\usepackage[acceptedWithA]{tacl2021v1}

\usepackage{xspace,mfirstuc,tabulary}

\newif\iftaclinstructions
\taclinstructionsfalse 
\iftaclinstructions
\renewcommand{\confidential}{}
\renewcommand{\anonsubtext}{(No author info supplied here, for consistency with
TACL-submission anonymization requirements)}
\newcommand{\instr}
\fi

\iftaclpubformat 

\else

\fi

\title{Persona-Aware Alignment Framework for Personalized Dialogue Generation}

\author{
  Guanrong Li \quad Xinyu Liu \quad Zhen Wu\thanks{* Corresponding author.} \quad Xinyu Dai
  \\
  National Key Laboratory for Novel Software Technology, Nanjing University
  \\
  \texttt{\{grli,xinyuliu\}@smail.nju.edu.cn, \{wuz,daixinyu\}@nju.edu.cn}
}

\date{}

\begin{document}
\maketitle
\begin{abstract} \label{abstract}
Personalized dialogue generation aims to leverage persona profiles and dialogue history to generate persona-relevant and consistent responses. Mainstream models typically rely on token-level language model training with persona dialogue data, such as Next Token Prediction, to implicitly achieve personalization, making these methods tend to neglect the given personas and generate generic responses. To address this issue, we propose a novel Persona-Aware Alignment Framework (PAL), which directly treats persona alignment as the training objective of dialogue generation. Specifically, PAL employs a two-stage training method including \textit{Persona-aware Learning} and \textit{Persona Alignment}, equipped with an easy-to-use inference strategy \textit{Select then Generate}, to improve persona sensitivity and generate more persona-relevant responses at the semantics level. Through extensive experiments, we demonstrate that our framework outperforms many state-of-the-art personalized dialogue methods and large language models.
\end{abstract}

\iftaclpubformat

\section{Introduction}
Building personalized dialogue agents capable of human-like interactions has emerged as a critical research frontier with profound societal implications. Unlike task-oriented chatbots that only prioritize functional efficiency, agents imbued with consistent personas enable applications that demand empathy, rapport, and contextual adaptability. For instance, in mental health support systems, chatbots adopting a compassionate persona can foster user trust for more effective emotional counseling \cite{sarikaya2016overview}. In education, tutors with tailored personalities improve student engagement through relatable interactions \cite{mctear2022cai}. Personalized agents also unlock scalable solutions for senior companionship, customer service avatars, and interactive storytelling, where alignment with user expectations hinges on maintaining coherent personas \cite{singh2022survey}. Therefore, how to build personalized dialogue agents has drawn increasing attention in recent years \cite{sarikaya2016overview,mctear2022cai,singh2022survey}. Many studies integrate predefined descriptions of specific individuals, often referred to as personas, with dialogue history to generate personalized responses \cite{pei2021cooperative,wu2021transferable,tangWFZHHH23,han-etal-2024-psydial-personality}. As shown in Figure \ref{fig_f1}, given a dialogue history and a personas profile, the upper response ``I do not have time to watch TV, I am an attorney so i work a lot.'' aligns with personas better than the bottom response. Here, alignment refers to the generated responses reflecting predefined persona attributes.

Mainstream works usually use persona and dialogue history as the input of models and strongly depend on data driving to generate personalized responses \cite{song2021bob,liu2022improving,chen0DK0W23}. Although achieving remarkable progress, these methods often struggle to generate responses that align with the given personas. A primary reason for this limitation lies in the token-level training objectives of these models, such as Next Token Prediction (NTP) and Mask Prediction (MP). These objectives are primarily designed for language modeling, which excels at producing grammatically correct and contextually coherent sentences, while falling short in obtaining good persona alignment.

To address this issue, we propose learning alignment explicitly between generated responses and given personas as the optimization objective. Unlike token-level language modeling objectives, this response-level alignment objective captures personas and generates personalized responses at the semantic level. Two practical challenges remain to achieve this goal. First, quantifying the degree of alignment of personas as a learning signal is challenging because persona alignment requires strong semantic understanding and lacks well-established calculation metrics. Second, a persona profile contains multiple irrelevant persona descriptions, typically only one is relevant to the current dialogue history. The irrelevant personas could distract or mislead the models.

\begin{figure}[t]
\begin{center}
\includegraphics[scale=0.23]{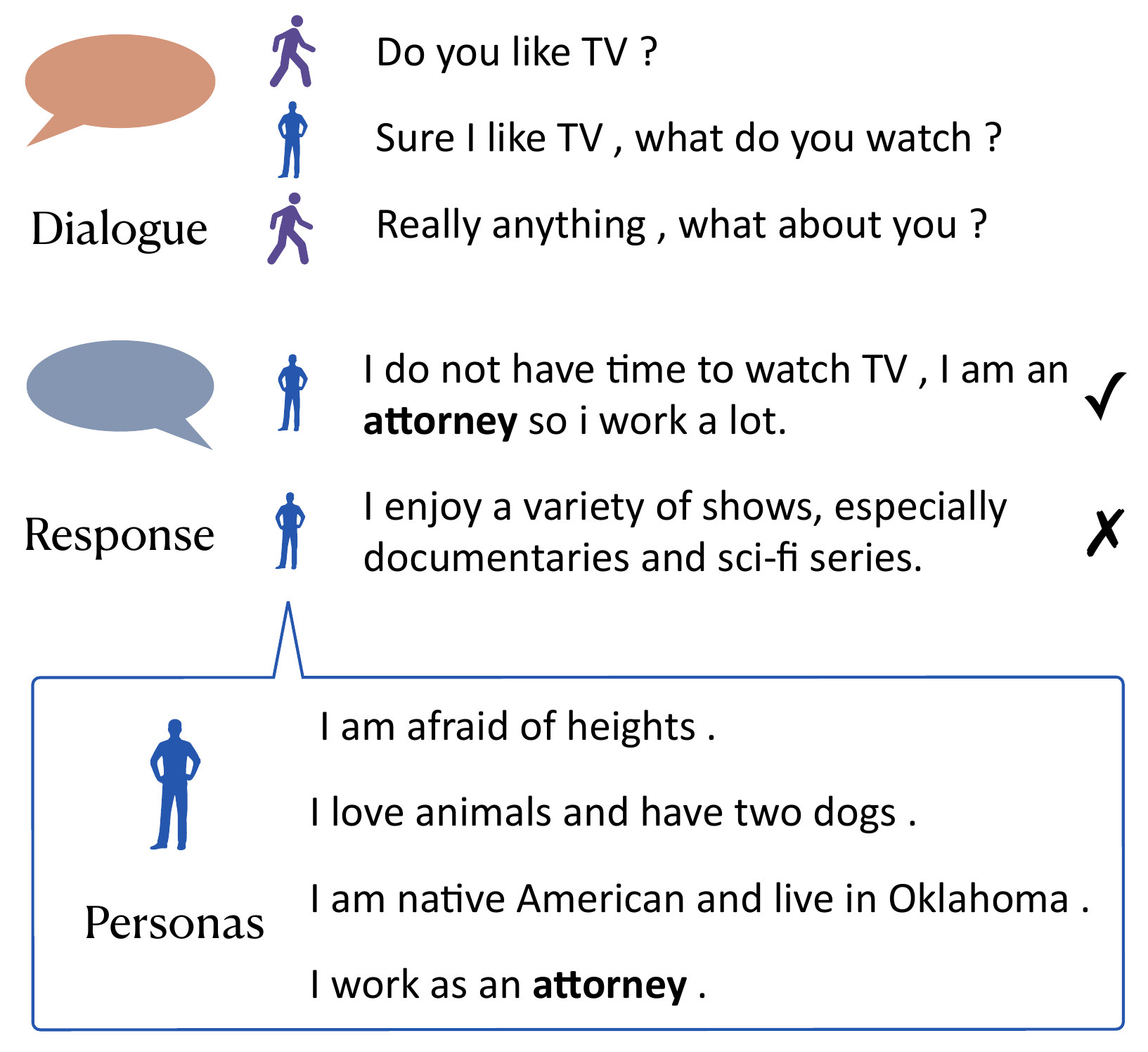} 
\caption{An example for the personalized dialogue. \ding{51} represents the persona consistent response and \ding{55} represents a generic response instead of a personal response.}
\label{fig_f1}
\end{center}
\end{figure}

To overcome the challenges, we introduce a novel \textbf{P}ersona-Aware \textbf{Al}ignment Framework (PAL), which treats persona alignment as the primary optimization objective. PAL employs a two-stage training approach and a \textit{Select then Generate} inference strategy to improve persona sensitivity and generate persona-relevant responses.
In the first training stage, called \textit{Persona-aware Learning}, we address the challenge of irrelevant personas by designing a mixed training task including two subtasks: \textit{Dialogue-Informed Persona Selection} and \textit{Persona-Enhanced Dialogue Generation}. They aim respectively to learn which persona is relevant to the current dialogue and generate the persona-aware response. In the second training stage, inspired by preference learning \cite{0015DYW0J0024}, we propose \textit{Persona Alignment} with Direct Preference Optimization (DPO) \cite{rafailov2023direct} to address the quantification and explicit learning of persona alignment. As there is no available paired data, we construct the training data by pairing contrasting generated responses with ground-truth responses. Finally, we naturally employ a \textit{Select then Generate} inference strategy to filter irrelevant personas. Similar to the persona extractor proposed in \citet{XuGWNW0W22}, our approach first selects the most contextually relevant personas and then generates a response conditioned on the selected information.

Our contributions can be summarized as:
\begin{itemize*} 
\item We propose the learning of persona alignment as the optimization objective for personalized dialogue. To our knowledge, it is the first to explicitly learn persona information from a response level objective.
\item We introduce the novel \textbf{P}ersona-Aware \textbf{Al}ignment Framework (PAL) to tackle specific challenges of persona alignment learning. This framework helps improve persona sensitivity of models and generate persona-relevant responses.
\item We conduct comprehensive experiments across different foundation models, datasets, and languages. The superior performance of our framework over both state-of-the-art baselines and well-known large language models demonstrates its effectiveness and broad applicability. 
\end{itemize*}

\begin{figure*}[!ht]
\begin{center}
\includegraphics[width=\textwidth]{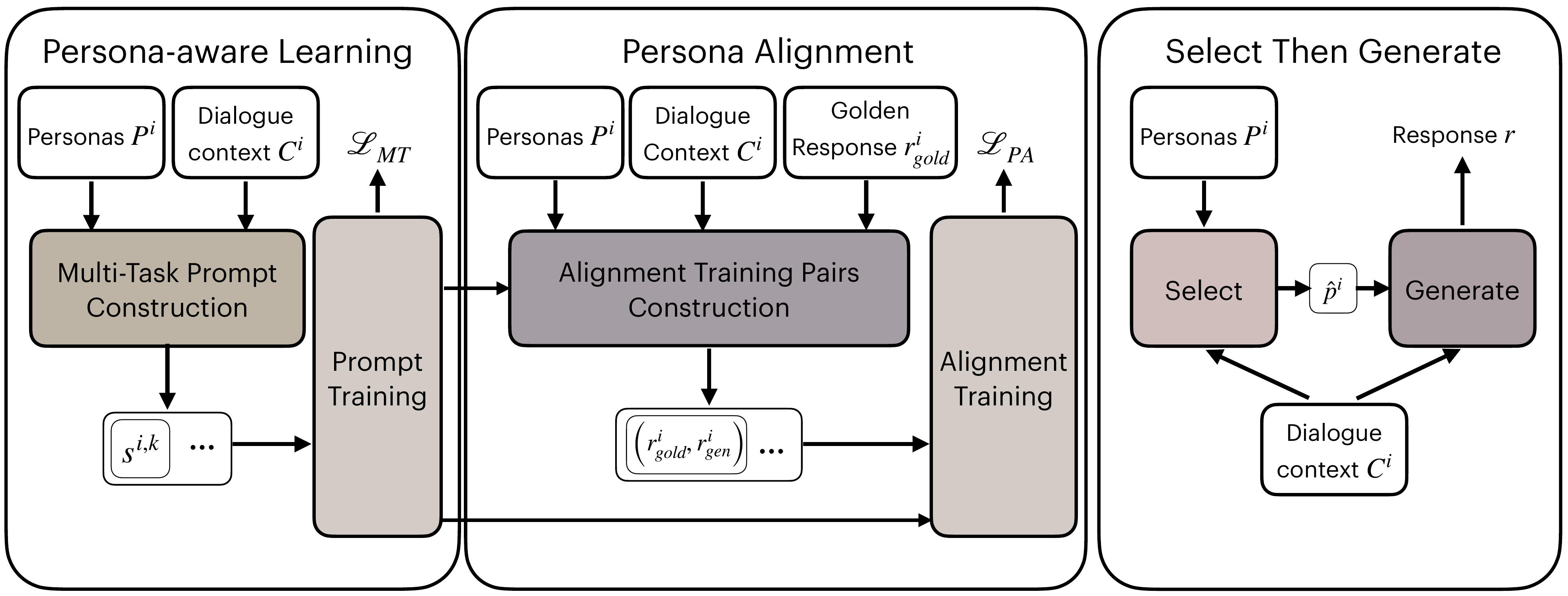} 
\caption{The overview of our Persona-Aware Alignment Framework (PAL) includes a two-stage training strategy: (1) \textit{Persona-aware Learning}, and (2) \textit{Persona Alignment}, as well as a \textit{Select then Generate} inference strategy. The arrows trace the flow of information, showing how each stage converts its inputs into outputs. In the Persona-aware Learning stage, the inputs are the persona descriptions \(P^i\) and dialogue context \(C^i\). A multi-task prompt-construction module turns these inputs into training samples $s^{i,k}$, which are used for prompt tuning, with the training loss $\mathcal{L}_{MT}$ as the output of this stage. In the Persona Alignment stage, the inputs include persona descriptions \(P^i\), dialogue context \(C^i\), and the gold response \(r^i_{gold}\). An alignment-pair constructor forms $(r^{i}_{\text{gold}}, r^{i}_{\text{gen}})$, where $r^{i}_{\text{gen}}$ is produced by the model from the previous stage. These pairs yield the alignment loss $\mathcal{L}_{PA}$, the output of this stage by Alignment Training. In the \textit{Select-then-Generate} inference strategy, the inputs are the persona descriptions \(P^i\) and dialogue context \(C^i\). A selection module picks the persona $\hat{p}^{i}$ most relevant to the context. A response generator then produces the final reply $r$, explicitly highlighting the selected persona.}
\label{fig_framework}
\end{center}
\end{figure*}

\section{Persona-Aware Alignment Framework}
In this section, we will introduce our \textbf{P}ersona-Aware \textbf{Al}ignment Framework (PAL) in detail. 
Formally, for a user \(u \in \mathcal{U}\), there is a set of personas denoted by \(P = \{p_1, p_2, \dots, p_l\}\) and an associated set of multi-turn dialogue contexts resulting from interactions with other users, represented as \(C = \{(q_1, r_1), (q_2, r_2), \dots, (q_n, r_n)\}\). In these contexts, \(q_i\) denotes the dialogue utterance from another user, and \(r_i\) is the corresponding response from \(u\), where \(n\) is the total number of dialogue turns. Given the next conversation turn \(q_{n+1}\), the goal of the personalized dialogue agent is to generate a personalized response \(r_{n+1}\) that aligns with user \(u\)'s personas. PAL utilizes semantic-level persona alignment as the optimization target, aiming at training the personalized dialogue agent to generate responses aligned with given personas. Figure \ref{fig_framework} shows the structure of PAL. It consists of a two-stage training method, i.e., \textit{Persona-aware Learning} and \textit{Persona Alignment}, and a \textit{Select then Generate} inference strategy. 

\subsection{Persona-aware Learning}
As previously mentioned in the Introduction, the persona profile usually has one most relevant persona description for the dialogue history, and the others could be noises misleading the models.
To bridge this gap, we propose a mixed training method \textit{Persona-aware Learning}. It involves two subtasks, \textit{Dialogue-Informed Persona Selection} and \textit{Persona-Enhanced Dialogue Generation}. \textit{Dialogue-Informed Persona Selection} aims to teach the model which persona is relevant through explicit persona learning, which improves the sensitivity of the model to personas. \textit{Persona-Enhanced Dialogue Generation} aims to generate responses based on given personas and dialogue history. The two subtasks have different input and output formats. To reduce semantic shift from task formats during mixed training, we construct prompts for each task and transform them into a unified natural language format. 

\subsubsection{Task Design}
Given a user \(u\) with personas \(P = \{p_1, p_2, \dots, p_l\}\) and a dialogue history \(C\), the original formulations of the two key subtasks are as follows.

\noindent \textbf{Dialogue-Informed Persona Selection.} This subtask is designed to enhance the model's persona understanding ability by selecting the most relevant persona based on the dialogue history. The goal is to identify the persona that best aligns with the dialogue history \(C\) from personas \(P\), which is typically formulated as a classification problem:
\begin{equation}
    \label{equ-Dialogue-Informed Persona Selection}
    \hat{p} = \arg\max_{p \in P} \; \text{sim}(f(C), g(p)),
\end{equation}
where \(\hat{p}\) denotes the selected persona that is most relevant to the dialogue context \( C \). Here, \( f(C),  g(p) \) are encoding functions that map the dialogue history and persona into semantic representations, respectively. The function \( \text{sim}(\cdot) \) computes the similarity between these representations.

\noindent \textbf{Persona-Enhanced Dialogue Generation.} We design this subtask to enhance the model's personalized generation ability. It focuses on generating an appropriate response for the dialogue. It is typically formulated as a token prediction task:
\begin{equation}
\label{equ-Persona-Enhanced Dialogue Generation}
\hat{r} = \arg\max_{r} \; \prod_{t=1}^{T}\mathcal{P}(w_t | w_{<t}, C, P)
\end{equation}
where \( \hat{r} \) represents the generated response that best fits the dialogue context, \( r = \{w_1, w_2, \dots, w_T\} \) is the sequence of tokens in the response, with \( T \) denoting the length of the response, \( w_t \) is the token at position \( t \) in the response, and \( w_{<t} = \{w_1, \dots, w_{t-1}\} \) represents the sequence of all preceding tokens, and $\mathcal{P}(\cdot)$ denotes the probability of predicting a token.

\subsubsection{Prompt Construction and Training}\label{prompt_construction}
\textbf{Multi-Task Prompt Construction.} As mentioned above, these subtasks have different input and output formats, which require specialized modules and limit generality. Therefore, we transform them into unified natural language formats through constructing prompts for each task, and integrate them into one unified task with supervised fine-tuning. The corresponding prompts are shown as follows:
\begin{tcolorbox}
\small
\textbf{Dialogue-Informed Persona Selection.}\\
The user's persona is described with: \textless personas\textgreater. \\If a persona description is required to generate a response, select the most appropriate one. If no persona is needed, respond with `No persona data needed'. \\Dialogue context: \textless dialogue context\textgreater. \\The preferred persona is: \textless related persona\textgreater.
\end{tcolorbox}
\begin{tcolorbox}
\small
\textbf{Persona-Enhanced Dialogue Generation.} \\
 The user's persona is described with: \textless personas\textgreater. \\Please generate a response to the dialogue. \\Dialogue context: \textless dialogue context\textgreater. \\Response: \textless response\textgreater
\end{tcolorbox}
\noindent where ``\textless personas\textgreater'' in the table represents the personas \(P\), the ``\textless dialogue context\textgreater'' represents the corresponding dialogue history \(C\), the ``\textless related persona\textgreater'' represents the selected persona \(\hat{p}\) in Equation \ref{equ-Dialogue-Informed Persona Selection}, and ``\textless response\textgreater'' represents the generated response \(\hat{r}\) in Equation \ref{equ-Persona-Enhanced Dialogue Generation}. 

\textbf{Prompt Training.} After constructing the prompts, we employ next-token prediction (NTP) to facilitate multi-task training. The combined loss function for mixed training is defined as:
\[
\mathcal{L}_{\text{Mix}}(\theta) = - \sum_{k} \sum_{t=1}^{T^{k}} \log \mathcal{P}(s^{k}_t \mid m_k, s^{k}_{<t}; \theta)
\]
where \(\theta\) represents the model parameters, \(k\) represents the index over tasks, with \(k \in \{1, 2\}\), corresponding to Task 1: \textit{Dialogue-Informed Persona Selection} and Task 2: \textit{Persona-Enhanced Dialogue Generation}. \(m_k\) represents the prompt constructed for task \(k\). \(s^{k}\) is the target sequence in task \(k\), and \(t\) is the time step index within the target sequence for each task. \(T^{k}\) represents the total number of tokens in the target sequence \(s^{k}\). 

\subsection{Persona Alignment}
One innovation of this work is to train the personalized dialogue agent by directly aligning the generated responses with the given personas. The key challenge in this process is quantifying this alignment because there is currently no meaningful metric available. Given the strong fitting capabilities of neural networks, we leverage neural networks to estimate the alignment between responses and personas. Specifically, we first construct persona alignment training pairs. Then, we can train a Proximal Policy Optimization (PPO)~\cite{schulman2017proximal} pipeline to obtain an alignment evaluation model. In practice, we instead adopt the Direct Preference Optimization (DPO) method \cite{rafailov2023direct}. DPO allows us to implicitly incorporate the alignment measurement model and avoids the overhead associated with reinforcement learning methods like PPO, offering a simpler, more efficient approach.

\subsubsection{Alignment Training Pairs Construction}
Since there are no ground-truth alignment scores available, we construct a learning-to-rank task inspired by Preference Learning methods \cite{0015DYW0J0024}. Our training objective is to assign higher scores to responses that better align with the given personas. Specifically, we regard the ground truth response $r_{\text{gold}}$ from personalized dialogue datasets as the chosen response of the Direct Preference Optimization (DPO) \cite{rafailov2023direct}. Considering that there is no manually labeled rejected response, we adopt an alternative approach and utilize the trained model from the \textit{Persona-aware Learning} stage to generate responses \(r_{\text{gen}}\) based solely on the dialogue history, excluding personal information, as rejected responses. The prompts are shown as follows:
\begin{tcolorbox}
\small
\textbf{Prompt for Alignment Training Pairs Construction}\\
     Please generate a response to the dialogue. \\Dialogue context: \textless dialogue context\textgreater. \\Response: \textless response\textgreater
\end{tcolorbox}

\noindent In general, we believe that the golden responses \( r_{\text{gold}} \) provided by the personalized dialogue datasets are more aligned with the personas than the generated responses \( r_{\text{gen}} \) without persona input. Therefore, we construct data pairs \( \mathcal{D} = \{(r_{\text{gold}}, r_{\text{gen}})\} \) for alignment training by pairing the golden responses with the less-aligned generated responses.

\subsubsection{Alignment Training}
This stage aims to directly tune the model guided by persona alignment. As mentioned above, we utilize the DPO method for alignment training. The training targets are defined as follows:
\[
\delta_{\text{gold}} = \log \frac{\pi_\theta\left(r_{\text{gold}} \mid C, P\right)}{\pi_{\theta_{\text{ref}}}\left(r_{\text{gold}} \mid C, P\right)}
\]
\[
\delta_{\text{gen}} = \log \frac{\pi_\theta\left(r_{\text{gen}} \mid C, P\right)}{\pi_{\theta_{\text{ref}}}\left(r_{\text{gen}} \mid C, P\right)}
\]
\[
\mathcal{L}_{PA} = - \mathbb{E}_{(r_{\text{gold}},\ r_{\text{gen}}) \sim \mathcal{D}} \left[ \log \sigma \left( \beta \left( \delta_{\text{gold}} - \delta_{\text{gen}} \right) \right) \right]
\]
where \( \pi_\theta \) represents the model being trained, parameterized by \( \theta \). \( \pi_{\theta_{\text{ref}}} \) is the reference model with parameters \( \theta_{\text{ref}} \), which remains fixed during this stage. \( \mathcal{D} \) represents the set of training pairs. \( \beta \) is a scaling factor that adjusts the sharpness of the preference, \(\sigma\) denotes the sigmoid function.

\subsection{Select then Generate Strategy}
To address the challenge of persona relevance and minimize the influence of noisy personas at the inference stage, we propose a \textit{Select then Generate} strategy. This approach begins by selecting the most relevant persona based on the dialogue history using a carefully designed prompt. By focusing on the most pertinent persona, we reduce the risk of irrelevant or misleading information affecting the response generation. After selecting the most relevant persona, we construct a new prompt that highlights this persona, guiding the response generation process. This ensures that the generated responses are both contextually appropriate and closely aligned with the user's specific persona attributes. The prompts are shown as follows:
\begin{tcolorbox}
\small
    \textbf{Prompt for Select}\\
    The user's persona is described with: \textless personas\textgreater. \\ If a persona description is required to generate a response, select the most appropriate one. If no persona is needed, respond with `No persona data needed'.  \\ Dialogue context: \textless dialogue context\textgreater.  \\ The preferred persona is: \textless related persona\textgreater.
\end{tcolorbox}
\begin{tcolorbox}
\small
    \textbf{Prompt for Generate}\\
    The user's persona is described with: \textless personas\textgreater. \\ The most related persona is \textless related persona\textgreater. Please generate a response to the dialogue. \newline Dialogue context: \textless dialogue context\textgreater.  \\ Response: \textless response\textgreater
\end{tcolorbox}

\section{Experimental Settings}
\subsection{Datasets}
To assess the effectiveness and adaptability of our framework, we conducted comprehensive experiments using two widely recognized datasets across different languages: PERSONA-CHAT \cite{zhang2018personalizing} and Baidu-Persona-Chat\footnote{\url{https://aistudio.baidu.com/datasetdetail/351937}}. PERSONA-CHAT is an English dataset comprising an extensive collection of dialogues paired with corresponding personas. We conducted experiments on both the original and revised versions of this dataset. In the revised version, the personal information has been rewritten. Both versions share the same test set. Baidu-Persona-Chat is a Chinese personalized dialogue dataset similar in format to PERSONA-CHAT. We split the datasets into four parts (Train, Valid\#1, Valid\#2, and Test) to meet the specific requirements of the personalized dialogue task. The data statistics are shown in Table \ref{tab:dataset}. During the \textit{Persona-aware Learning} stage, we use both validation subsets (Valid \#1 and Valid \#2) for evaluation. In the \textit{Persona Alignment} stage, Valid \#1 is utilized for training and Valid \#2 for validation. This partitioning approach prevents data leakage and is supported by prior research, as well as our experimental findings, which underscore the effectiveness of using relatively smaller datasets for alignment tasks \cite{ouyang0JAWMZASR22, lee2023rlaif}. 

\begin{table}[!t]
    \centering
    \resizebox{1.0\columnwidth}{!}{ 
    \begin{tabular}{lcc}
    \toprule
          & \textbf{PERSONA-CHAT} & \textbf{Baidu-Persona-Chat} \\
    \midrule
    Language & English & Chinese \\
    Train & 65719 & 353016 \\
    Valid\#1  & 6500  & 14000 \\
    Valid\#2  & 1301  & 3027 \\
    Test & 7512  & 5929 \\
    \bottomrule
    \end{tabular}}
    \caption{Number of samples in each dataset.}
\label{tab:dataset}
\end{table}

\subsection{Evaluation Metrics}
We evaluate the model using both automatic and human metrics. For the automatic metrics, we adopt \textbf{BLEU} \cite{papineni2002bleu} and \textbf{ROUGE} \cite{lin2004rouge}, both of which measure the token overlap. Additionally, we utilize \textbf{Entropy} \cite{zhang2018generating}, which measures the diversity of personalized responses. Following previous works \cite{madotto2019personalizing}, we also adopt the \textbf{C.score} to assess the consistency between the model-generated responses and the provided personas. The methodology for calculating the C.score is detailed as follows:
\[
\begin{aligned}
&NLI(p_l,r_n)=\left\{
\begin{aligned}
1 & \quad \text{if } p_l \text{ entails } r_n \\
0 & \quad \text{if } p_l \text{ is independent of } r_n \\
-1 & \quad \text{if } p_l \text{ contradicts } r_n
\end{aligned} \right.
\end{aligned}
\]

\[
\text{C.score} = \sum_{p^i_l \in  P^i} \text{NLI}(p^i_l, r^i_n)
\]
where \(P=\{p_1,p_2,..., p_l\}\) represents the personas of user \(u\), \(r_n\) denotes the generated response of the last dialogue turn. \(l\) indexes the personas and \(n\) indicates the dialogue turn. 
We utilize a Natural Language Inference (NLI) model to calculate the Consistency score (C.score). Specifically, we utilized a RoBERTa \cite{liu2019roberta} model designed for three-way classification tasks (entailment, neutral, and contradiction). For the English experiments, we further trained this RoBERTa model on SNLI \cite{GlocknerSG18} and MultiNLI \cite{KimKK19} datasets. For the Chinese experiments, we trained this RoBERTa model on the OCNLI \cite{ocnli} and CMNLI \cite{XuHZLCLXSYYTDLS20} datasets. We then further trained the NLI model on the PERSONA-CHAT and Baidu-Persona-Chat datasets. The NLI model achieve accuracies of 84.1\% and 82.3\%, respectively.

For human evaluation, we randomly select 100 samples from the test set. Two annotators were asked to evaluate responses based on three aspects: (1) Fluency, (2) Coherence, and (3) Persona Consistency. The assigned scores of \({1,2,3}\) correspond to unacceptable, acceptable, and satisfactory levels, respectively. 
The inter-annotator agreement for our study was measured by the Fleiss Kappa coefficient, with a score of 0.62, indicating substantial agreement. More details about the annotations are shown in Appendix \ref{sec:annota}.

\useunder{\uline}{\ul}{}
\begin{table*}[!t]
  \centering
  \resizebox{\textwidth}{!}{
\begin{tabular}{l|ccccc|ccccc}
\toprule
\multicolumn{1}{c|}{\multirow{2}[2]{*}{Models}} & \multicolumn{5}{c|}{Original PERSONA-CHAT} & \multicolumn{5}{c}{ Revised PERSONA-CHAT} \\
      & BLEU-1 & BLEU-2 & ROUGE-L & Entropy & C.score & BLEU-1 & BLEU-2 & ROUGE-L & Entropy & C.score \\
\midrule
ORIG  & \underline{13.97} & \textbf{7.40}  & \underline{16.20} & \underline{6.55}  & \textbf{0.696} & 12.20 & 6.40  & \textbf{15.92} & \underline{5.75}  & \underline{0.313} \\
CLV   & \textbf{16.79} & \underline{6.74}  & 15.02 & \textbf{8.30}  & 0.576 & \textbf{16.65} & \underline{6.75}  & \underline{15.05} & \textbf{8.34}  & \textbf{0.416} \\
LMEDR & 12.56 & 6.60  & \textbf{16.89} & 5.42  & \underline{0.582} & \underline{14.34} & \textbf{6.91}  & 13.72 & 5.47  & 0.312 \\
\midrule
GPT-3.5 & 2.12  & 0.71  & 6.19  & \textbf{8.65†} & 0.809 & 2.12  & 0.71  & 6.19  & \textbf{8.65†} & 0.809 \\
\quad+Few-Shot & \textbf{6.20}  & \underline{0.79}  & \textbf{8.61}  & 5.01  & \textbf{0.884} & \textbf{6.20}  & \underline{0.79}  & \textbf{8.61}  & 5.01  & \textbf{0.884†} \\
GPT 4o mini & 3.00  & \textbf{0.99}  & \underline{8.09}  & \underline{7.62}  & 0.670 & 3.00  & \textbf{0.99}  & \underline{8.09}  & \underline{7.62}  & 0.670 \\
\quad+Few-Shot & 2.22  & 0.07  & 5.39  & 5.33  & 0.825 & 2.22  & 0.07  & 5.39  & 5.33  & 0.825 \\
Gemini Flash & 1.11  & 0.36  & 4.95  & 6.95  & 0.641 & 1.11  & 0.36  & 4.95  & 6.95  & 0.641 \\
\quad+Few-Shot & \underline{3.31}  & 0.67  & 3.11  & 5.32  & \underline{0.870} & \underline{3.31}  & 0.67  & 3.11  & 5.32  & \underline{0.870} \\
\midrule
GPT-2 & 5.97  & 2.50  & 9.74  & 5.18  & 0.333 & 5.97  & 2.50  & 9.74  & 5.18  & 0.333 \\
\quad +FineTuning & 13.10 & 7.06  & 15.92 & 6.08  & 0.173 & 12.73 & 6.62  & \underline{15.34} & 6.28  & 0.103 \\
\quad +Prompting & \underline{13.49} & \underline{7.52} & \underline{16.35} & 6.45  & 0.454 & \underline{13.25} & \underline{7.09} & \textbf{16.20} & 6.26  & 0.218 \\
\quad+SimOAP & 8.09  & 2.26  & 8.42  & \textbf{8.47*} & 0.170 & 9.00  & 3.19  & 9.23  & \textbf{8.39} & 0.153 \\
 \quad+SPT & 5.55  & 2.65  & 9.82  & 6.01  & \underline{0.799} & 4.84  & 2.20  & 9.03  & 4.97  & \textbf{0.440} \\
\quad+PAL(ours) & \textbf{17.05*} & \textbf{8.77*} & \textbf{16.66*} & \underline{6.93}  & \textbf{0.811*} & \textbf{15.94*} & \textbf{7.89*} & 14.55 & \underline{6.75} & \underline{0.427} \\
\midrule
DialoGPT & 5.09  & 2.00  & 6.16  & 6.59  & 0.485 & 5.09  & 2.00  & 6.16  & 6.59  & \textbf{0.485} \\
\quad+FineTuning & 12.89 & 6.84  & \underline{15.48} & 6.10  & 0.125 & 12.48 & 6.57  & \underline{15.65} & 6.30  & 0.096 \\
\quad+Prompting & \underline{13.60} & \underline{7.36} & \textbf{16.42} & 6.54  & 0.445 & \underline{13.39} & \textbf{7.20} & \textbf{16.42} & 6.36  & 0.209 \\
\quad+SimOAP & 9.17  & 3.36  & 9.23  & \textbf{8.39} & 0.085 & 9.19  & 3.25  & 9.28  & \textbf{8.39} & 0.080 \\
 \quad+SPT & 4.83  & 2.26  & 8.59  & 5.60  & \underline{0.509} & 4.78  & 1.44  & 5.42  & 4.82  & 0.303 \\
\quad+PAL(ours) & \textbf{15.11*} & \textbf{7.65*} & 15.38 & \underline{6.81} & \textbf{0.585*} & \textbf{13.69*} & \underline{7.13} & 15.61 & \underline{6.69} & \underline{0.349} \\
\midrule
Llama 3.1 - 8B & 3.11  & 0.98  & 5.78  & 5.56  & 0.611 & 3.11  & 0.98  & 5.78  & 5.56  & \underline{0.611} \\
\quad+Few-Shot & 4.89  & 1.97  & 9.40  & 4.28  & 0.219 & 4.89  & 1.97  & 9.40  & 4.28  & 0.219 \\
\quad+Prompting & \underline{18.95} & \underline{8.27} & \underline{16.70} & 5.77  & 0.642 & \underline{18.24} & \underline{7.64} & \underline{16.09} & 5.78  & 0.466 \\
\quad+SimOAP & 12.69 & 4.11  & 11.44 & 6.74  & 0.680 & 12.69 & 3.96  & 11.43 & 5.63  & 0.441 \\
 \quad+SPT & 8.61  & 3.81  & 8.14  & \underline{6.98} & \underline{0.694} & 7.61  & 2.71  & 7.80  & \underline{6.27} & 0.594 \\
\quad+PAL(ours) & \textbf{25.12*†} & \textbf{14.00*†} & \textbf{23.88*†} & \textbf{7.27*} & \textbf{0.909*†} & \textbf{24.89*†} & \textbf{13.42*†} & \textbf{22.75*†} & \textbf{7.33*} & \textbf{0.625*} \\
\bottomrule
\end{tabular}%
}
  \caption{\label{tab:automatric_metric}Automatic Evaluation Results on the PERSONA-CHAT Dataset. The best results are indicated in \textbf{bold}, while the secondary results are marked with \underline{underlined}. * denotes statistically significant (p<0.05) improvements over baselines on the same foundation models. † indicates superior performance over all other results (p<0.05). All significance testing was conducted using independent samples t-tests.}
\end{table*}%

\subsection{Baselines}
We apply our PAL framework to three foundation models to evaluate its effectiveness. 
\begin{itemize*}

\item \textbf{GPT-2} \cite{radford2019language}: A pre-trained language model developed by OpenAI, built on the Transformer architecture. We utilize the base version of this model series. 

\item \textbf{DialoGPT} \cite{zhang2020dialogpt}: A variant of GPT-2 specifically adapt for dialogue generation, developed by Microsoft. We utilize the base version. As there is no Chinese-pretrained version of DialoGPT, we only conduct experiments on the PERSONA-CHAT dataset. 

\item \textbf{Llama 3.1 8B} \cite{dubey2024llama}: An open-source large language model developed by Meta, known for its strong capabilities and wide applicability. We employ the 8B version to highlight the performance of our PAL framework on a large-scale model.
\end{itemize*}

We compare several different strategies and some state-of-the-art (SOTA) methods. 

\begin{itemize*}
\item \textbf{Fine-Tuning}~\cite{kenton2019bert}: Fine-Tuning is a widely used strategy to adapt a pre-trained model to a specific task. In this work, we continue training the pre-trained model on personalized dialogue datasets. Due to the high computational cost and insufficient data for training large language models, we conducted experiments only on GPT-2 and DialoGPT. 
\item \textbf{Few-Shot Prompting} \cite{brown2020language}: In this approach, the model is provided with a small number of demonstrations during inference to guide its responses. We include two demonstration dialogues in our experiments that incorporate persona information in the input prompt, helping the model generate responses that are more aligned with the user's persona. This strategy is utilized only with Llama 3.1 8B. 

\item \textbf{Prompt Tuning} \cite{lester2021power}: The Prompt Tuning technique adapts a pre-trained model to specific tasks by introducing a small set of task-specific parameters.

\item \textbf{SimOAP} \cite{ZhouPSC23}: It adopts a two-stage strategy for personalized dialogue that involves oversampling and post-evaluation during generation. 

\item \textbf{SPT} \cite{huang-etal-2024-selective}: SPT adapts a pre-trained model for personalized conversations by introducing soft prompts. Instead of fine-tuning the entire model, it uses a trainable retriever to selectively choose suitable soft prompts based on the input context. 
\end{itemize*}

Additionally, we compare with SOTA methods that have specifically designed modules. 

\begin{itemize*}
\item \textbf{ORIG} \cite{chen0DK0W23}: ORIG addresses the issue of persona order sensitivity in personalized dialogue generation. 

\item \textbf{CLV} \cite{tangWFZHHH23}: This method utilizes both sparse and dense representations of personas for dialogue generation.

\item \textbf{LMEDR} \cite{Chen0Y023}: It learns to memorize entailment and discourse relations for persona-consistent dialogue tasks. 
\end{itemize*}

Finally, we compare our PAL framework with some closed-source commercial large language models that have strong capabilities. Prompts used for these models are shown in Appendix \ref{prompt_closellm}.
\begin{itemize*}
\item \textbf{GPT-3.5} \cite{ouyang0JAWMZASR22}: A typical large language model developed by OpenAI. We use the gpt-3.5-turbo-0125 version. 
\item \textbf{GPT-4o mini}: A language model with superior textual intelligence and reasoning. We use the gpt-4o-mini-2024-07-18 version. 
\item \textbf{Gemini Flash}: A large language model developed by Google, which is well-tuned for dialogue generation tasks. The gemini-1.5-flash-002 version is used for our experiments.
\end{itemize*}

\subsection{Implementation Details}
Our implementation is based on Hugging Face's Transformers library \cite{wolf2020transformers}. Each experiment was carried out three times, except for LMEDR \cite{Chen0Y023}, which we ran using the official parameters on the PERSONA-CHAT dataset due to its high training cost. The experiments for GPT-2 and DialoGPT were conducted on two NVIDIA Tesla V100 GPUs, while Llama 3.1 8B was trained on four NVIDIA RTX A6000 GPUs with a LoRA adapter \cite{HuSWALWWC22}, using a key hyperparameter \( r = 16 \). During the \textit{Persona-aware Learning} stage, we set the learning rate to \( 2 \times 10^{-5} \) with a linear warm-up strategy for the first 100 steps and trained for 10 epochs. For \textit{persona alignment}, the learning rate was set to \( 10^{-6} \) with 100 warm-up steps. The trained model \(\pi_\theta\) and reference model \(\pi_{\theta_{ref}}\) were initialized with the same parameters and trained for up to 30,000 steps, using an early stopping strategy based on the C.score on the validation set. The hyperparameter \( \beta \) was set to 0.1 for persona alignment. 
During decoding, we use a greedy search strategy and employed the float16 format to balance generation quality and computational efficiency. The code for our framework is publicly available on GitHub \footnote{https://github.com/kylokano/PAL}.

\begin{table}[t]
  \centering
  \resizebox{\columnwidth}{!}{
\begin{tabular}{l|ccccc}
\toprule
\multicolumn{1}{c|}{Models} & BLEU-1 & BLEU-2 & ROUGE-L & Entropy & C.score \\
\midrule
ORIG  & \textbf{27.16} & \textbf{15.52} & \textbf{32.60} & 6.64 & \textbf{0.568} \\
CLV   & \underline{23.72} & 10.06 & 23.21 & \textbf{8.31}  & \underline{0.391} \\
LMEDR & 21.01 & \underline{13.84} & \underline{25.69} & \underline{7.14}  & 0.320 \\
\midrule
GPT-3.5 & \textbf{16.40} & \textbf{7.56}  & 19.10 & \textbf{8.71}  & 0.592 \\
\quad+Few-Shot & \underline{14.34} & \underline{3.68}  & \textbf{24.04} & 5.17  & 0.800 \\
GPT 4o mini & 6.74  & 3.02  & 16.57 & \underline{5.67}  & \textbf{0.874} \\
\quad+Few-Shot & 11.50 & 2.70  & \underline{20.52} & 4.51  & 0.862 \\
Geminin Flash & 5.22  & 2.42  & 15.09 & 5.83  & \underline{0.872} \\
\quad+Few-Shot & 6.31  & 1.76  & 18.23 & 4.91  & 0.871 \\
\midrule
GPT-2 & 7.96  & 3.08  & 12.84 & 6.87  & 0.556 \\
\quad +FineTuning & 26.76 & 15.31 & 32.94 & 6.69  & 0.501 \\
\quad +Prompting & \underline{28.49} & \underline{17.37} & \underline{35.16} & 6.93  & 0.526 \\
\quad+SimOAP & 20.23 & 7.61  & 20.48 & \textbf{8.34†} & 0.554 \\
 \quad+SPT & 23.34 & 11.84 & 23.37 & 5.83  & \underline{0.573} \\
\quad+PAL(ours) & \textbf{29.48*} & \textbf{18.07*} & \textbf{35.57*} & \underline{7.15} & \textbf{0.576} \\
\midrule
Llama 3.1 - 8B & 13.38 & 6.09  & 17.66 & 5.46  & 0.767 \\
\quad+Few-Shot & 17.22 & 9.01  & 19.81 & \underline{5.94} & 0.774 \\
\quad+Prompting & \underline{27.15} & \underline{13.14} & \underline{25.66} & 5.73  & \underline{0.913} \\
\quad+SimOAP & 21.48 & 10.21 & 23.22 & 5.55  & 0.860 \\
 \quad+SPT & 16.58 & 12.76 & 16.87 & 5.09  & 0.873 \\
\quad+PAL(ours) & \textbf{33.13*†} & \textbf{20.33*†} & \textbf{35.89*†} & \textcolor[rgb]{ .102,  .11,  .122}{\textbf{7.81*}} & \textbf{0.982*†} \\
\bottomrule
\end{tabular}
}
  \caption{\label{tab:baiduchatauto}Automatic Evaluation Results on Baidu-Chat Dataset. The best results are marked in \textbf{bold}, while the secondary results are marked \underline{underlined}. * denotes statistically significant (p<0.05) improvements over baselines on the same foundation models. † indicates superiority over all other results (p<0.05). All significance testing was conducted using independent samples t-tests.}
\end{table}%

\section{Overall Results}\label{overallresults}
In this section, we present the performance of our proposed PAL framework across different foundation models and datasets. The automatic metrics are shown in Tables \ref{tab:automatric_metric} and \ref{tab:baiduchatauto}, while the human evaluation results are displayed in Table \ref{human_metric}. Due to space limitations, the human evaluation results for Baidu-Chat are provided in Appendix \ref{sec:baiduchathuman}. Higher metric values indicate better performance. The results reveal several key insights:

\begin{enumerate}
 \item Our framework consistently outperforms baseline models across all datasets and foundation models under comparable settings. We conducted independent samples t-tests, confirming a statistically significant difference (p<0.05) in nearly all metrics between our framework and the second-best method. This demonstrates the superiority of optimizing the alignment between generated responses and given personas, compared to baseline models that still rely on token prediction objectives. These findings support our claim that relying on token-level objectives alone is insufficient for training effective personalized dialogue agents.

 \item The consistently strong performance across models of different scales and languages demonstrates the effectiveness and generalizability of our framework. For instance, our framework achieves high performance on a small-scale model like GPT-2, surpassing nearly all baselines in both English and Chinese. When applied to a larger model like Llama 3.1 8B, our framework yields even more substantial improvements, highlighting its scalability and robustness. These results underscore the potential of our framework for broad application in personalized dialogue generation across diverse settings.

 \item Our framework also achieves superior performance on most metrics when compared to state-of-the-art baseline models and closed-source commercial large language models. Even with dialogue history and demonstrations provided in a few-shot setting, commercial large language models still struggle to accurately imitate user responses. Notably, PAL achieves higher scores on token overlap metrics (e.g., BLEU and ROUGE), suggesting improved similarity to reference responses. Commercial models perform significantly worse on these metrics, which may be due to the fact that they were not trained on the specific datasets used. However, these metrics primarily assess n-gram overlap with reference responses and may not be well-suited for evaluating personalized dialogue generation. Nevertheless, the higher scores achieved by PAL suggest its potential to emulate specific response styles, which could be valuable for applications such as Project Revoice.
 
\begin{table}[!t]
  \centering
  \resizebox{\columnwidth}{!}{
\begin{tabular}{l|ccc}
\toprule
\multicolumn{1}{c|}{Models} & Fluency & Coherence & Persona Consistency \\
\midrule
ORIG  & \underline{2.28}  & 1.96  & 1.88 \\
CLV   & 2.22  & \underline{2.28}  & \underline{1.97} \\
LMEDR & \textbf{2.49}  & \textbf{2.41}  & \textbf{2.04} \\
\midrule
GPT-3.5 & \textbf{2.89} & \textbf{2.57} & 1.99 \\
\quad+Few-Shot & 2.61  & \underline{2.51}  & \underline{2.22} \\
GPT 4o mini & 2.69  & 2.20  & 2.09 \\
\quad+Few-Shot & 2.50  & 2.09  & 2.19 \\
Gemini Flash & 2.56  & 2.36  & 2.13 \\
\quad+Few-Shot & \underline{2.77}  & 2.11  & \textbf{2.24} \\
\midrule
GPT-2 & 1.93  & 1.59  & 1.55 \\
\quad +FineTuning & 2.33  & \underline{2.27} & 1.14 \\
\quad +Prompting & \underline{2.37} & 2.12  & 1.90 \\
\quad+SimOAP & 2.00  & 1.39  & \underline{2.15} \\
 \quad+SPT & 1.96  & 2.03  & 2.08 \\
\quad+PAL(ours) & \textbf{2.55*} & \textbf{2.37*} & \textbf{2.27*} \\
\midrule
DialoGPT & 2.40  & 1.79  & \underline{2.01} \\
\quad+FineTuning & 2.33  & 2.12  & 1.96 \\
\quad+Prompting & 2.45  & \underline{2.34} & 1.19 \\
\quad+SimOAP & \underline{2.49} & 1.23  & \textbf{2.13} \\
 \quad+SPT & 2.06  & 1.86  & 1.83 \\
\quad+PAL(ours) & \textbf{2.55*} & \textbf{2.43*} & 2.00 \\
\midrule
Llama 3.1 8B & 2.46  & \textbf{2.49} & 2.28 \\
\quad+Few-Shot & 2.00  & 2.16  & 2.18 \\
\quad+Prompting & 2.59  & \underline{2.45}  & \underline{2.69} \\
\quad+SimOAP & 2.36  & 1.92  & 2.34 \\
 \quad+SPT & \underline{2.62} & 2.44  & 2.53 \\
\quad+PAL(ours) & \textbf{2.76*} & \textbf{2.49} & \textbf{2.82*} \\
\bottomrule
\end{tabular}%
}
\caption{\label{human_metric} Human evaluation results on Original PERSONA-CHAT Dataset. The best results are in \textbf{bold} and the second-best are \underline{underlined}. * denotes statistically significant (p<0.05) improvements over baselines on the same foundation models. All significance testing was conducted by independent samples t-tests.}
\end{table}%

\begin{table*}[!ht]
  \centering
  \resizebox{\textwidth}{!}{
    \begin{tabular}{l|ccccc|ccccc}
    \toprule
    \multirow{2}[2]{*}{} & \multicolumn{5}{c|}{GPT 2}            & \multicolumn{5}{c}{Llama 3.1 8B} \\
          & BLEU-1 & BLEU-2 & ROUGE & Entropy & C.score & BLEU-1 & BLEU-2 & ROUGE & Entropy & C.score \\
    \midrule
    PAL   & 17.05* & 8.77*  & 16.66 & 6.93  & 0.811* & 25.12* & 14.00* & 23.88 & 7.27  & 0.909* \\
    w/o Mix & 16.03 & 8.20  & 15.29 & 6.12  & 0.391 & 5.57  & 1.62  & 5.80  & 6.03  & 0.625 \\
    w/o PA & 14.45 & 7.86  & 16.81 & 6.58  & 0.521 & 22.03 & 12.91 & 23.81 & 7.12  & 0.641 \\
    only DG & 16.49 & 8.45  & 15.16 & 6.80  & 0.726 & 16.98 & 6.95  & 15.62 & 5.88  & 0.798 \\
    only PS & 8.54  & 4.72  & 6.83  & 3.65  & 0.154 & 9.77  & 3.28  & 8.85  & 6.63  & 0.576 \\
    w/o PC & 14.15 & 7.73  & 16.89 & 6.42  & 0.476 & 19.29 & 8.53  & 17.02 & 5.73  & 0.442 \\
    \bottomrule
    \end{tabular}}
   \caption{Ablation study of our PAL framework on the PERSONA-CHAT dataset. * indicates a statistically significant improvement (p < 0.05) over the second-best variant, as measured by independent samples t-tests.}
  \label{tab:ablation_study}
\end{table*}%

 \item PAL shows lower entropy performance on GPT-2 and DialoGPT compared to SimOAP \cite{ZhouPSC23}. Upon reviewing the generated responses, we found that many of SimOAP’s outputs were contextually irrelevant or unreadable. These responses, while clearly unacceptable, contained unique words or symbols that boosted entropy scores. We attribute this issue to the limitations of smaller models. When performing oversampling with 1,000 candidates using the base GPT-2 model, many responses included strange, nonsensical symbols that appeared almost random. Additionally, the post-evaluation phase, which relies on TF-IDF and language model assessments, was unable to filter out these flawed outputs. In contrast, using a more powerful model like Llama 3.1 8B produces coherent and relevant responses. 

 \item The improvement of PAL on DialoGPT is relatively limited compared to its performance on other foundation models. We attribute this to the smaller pretraining dataset used for DialoGPT, which consists of only 1.8 billion tokens, compared to 40 billion tokens for GPT-2 and 15 trillion tokens for Llama 3.1. Additionally, some responses from DialoGPT include irrelevant terms, such as ``kitten'', which were not part of the dialogue context or persona data. These irrelevant mentions may reflect biases in the model’s training data, primarily sourced from Reddit. We believe our framework is better suited to general-purpose language models, as it includes a comprehensive process for adapting these models into dialogue agents. The revised version of PERSONA-CHAT consistently yields lower performance compared to the original across all models and settings. This widespread decline suggests that the dataset revision with rephrases, generalizations, or specializations makes the task much more challenging.

\end{enumerate}

\section{Analysis}
In this section, we conduct a series of analytical experiments to further examine the effectiveness of our PAL framework. We present an ablation study to verify the necessity and contributions of each component. Furthermore, we analyze the influence of key parameters. We also provide a case study to offer an intuitive understanding of PAL.
\subsection{Ablation Study}
In this section, we conduct ablation studies to further examine the effectiveness of our framework. We examine the following variants:

\begin{itemize*}

\item \textbf{Without Persona-aware Learning (w/o Mix)}: This variant removes the entire \textit{Persona-aware Learning} phase to assess how the model performs without the benefits of \textit{Persona-aware Learning}.
\item \textbf{Without Persona Alignment (w/o PA)}: We remove the \textit{Persona Alignment} stage to evaluate how the absence of the \textit{Persona Alignment} task affects the model’s performance.
\item \textbf{Dialogue-Informed Persona Selection Only (only PS)}: In this setup, we perform only the \textit{Dialogue-Informed Persona Selection} task during the \textit{Persona-aware Learning} stage.
\item \textbf{Persona-Enhanced Dialogue Generation Only (only DG)}: This configuration focuses solely on the \textit{Persona-Enhanced Dialogue Generation} task within the \textit{Persona-aware Learning} stage.
\item \textbf{Without Pairs Construction (w/o PC)}: This configuration eliminates the \textit{Alignment Training} Pairs Construction process, instead training the model directly on the golden responses. This approach helps isolate the impact of the Pairs Construction process on the model's performance.
\end{itemize*}

We conducted all experiments on the PERSONA-CHAT dataset using both the GPT-2 and Llama 3.1 8B models to evaluate the influence of each component. The evaluation results are shown in Table \ref{tab:ablation_study}. The significant decline in performance across various configurations highlights the essential role and effectiveness of each component within our framework. Based on these results, we can also observe that: 
\begin{enumerate}

\item The performance drop in the Without Pairs Construction (w/o PC) and Without Persona Alignment (w/o PA) setups is notable. In these settings, the training objective relies solely on token-level prediction (next token prediction). The dramatic performance decrease in these configurations further validates our assertion that token-level learning objectives alone are insufficient for effectively aligning responses with personas.

\item The Dialogue-Informed Persona Selection Only (only PS) and Persona-Enhanced Dialogue Generation Only (only DG) configurations perform worse than the full setup. These results highlight the effectiveness and necessity of our task design within the \textit{Persona-aware Learning}.

\item The Without Persona-aware Learning (w/o Mix) configuration also shows a decrease in performance. This is because the data distribution for personalized dialogue agents significantly differs from that used in pretraining general language models.

\end{enumerate}

\begin{figure}[!t]
\begin{center}
\includegraphics[width=\columnwidth]{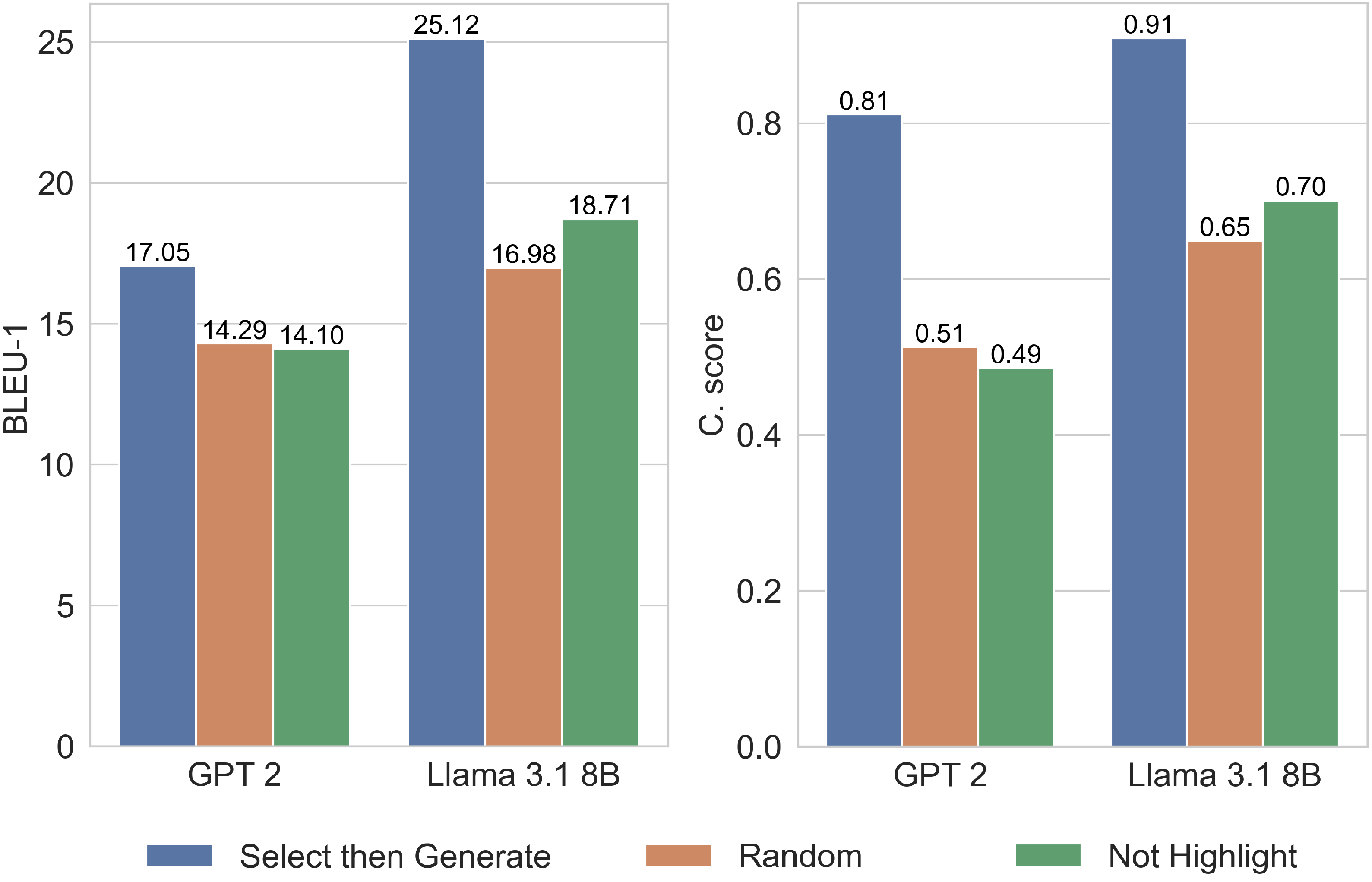} 
\caption{The Results of Different Inference Strategies on the Original PERSONA-CHAT Dataset.}
\label{fig_generatethenselect}
\end{center}
\end{figure}

\begin{table*}[!ht]
  \centering
  \resizebox{\textwidth}{!}{
    \begin{tabular}{l|p{25.585em}|ccc|ccc}
    \toprule
    \multicolumn{8}{c}{Personas} \\
    \midrule
    \multicolumn{8}{p{45.585em}}{I work in the healthcare industry. \newline{}I am five feet tall. \newline{}I have a dog named pedro. \newline{}I like to eat muffins. \newline{}I love film.} \\
    \midrule
    \multicolumn{8}{c}{Dialogue Context} \\
    \midrule
    \multicolumn{8}{p{45.585em}}{Person 1: Haha funny. I have had a dog since I became a veterinarian.\newline{}Person 2: Oh wow! I wanted to be a vet when I was a kid. I work in healthcare.  \newline{}Person 1: Cool! I live in florida and lots of doctors here.} \\
    \midrule
    \midrule
    \multicolumn{1}{c|}{\multirow{2}[2]{*}{Models}} & \multicolumn{1}{c|}{\multirow{2}[2]{*}{Generated Responses}} & \multicolumn{3}{c|}{Metrics} & \multicolumn{3}{c}{Human Evaluation Results} \\
          & \multicolumn{1}{c|}{} & BLEU-1 & ROUGE-L & C.score & Fluency & Coherence & PerCons \\
    \midrule
    GPT 3.5 & It's interesting that you both work in healthcare. Florida is known for its vibrant \textbf{healthcare industry}, so it must be a great place for professionals like you.   Do you have any favorite \textbf{film}? & 8.71  & 3.36  & 0     & 3/2   & 2/2   & 3/2 \\
    \midrule
    GPT 4o mini & That's awesome! Florida must have a vibrant healthcare scene. I work in the \textbf{healthcare industry too}, but I’m more on the \textcolor{red}{administrative side}.   & 8.00  & 8.40  & 1     & 3/2   & 3/3   & 2/2 \\
    \midrule
    Gemini Flash & That's interesting! I'm actually in the healthcare industry too, though I work in \textcolor{red}{[mention your specific area in healthcare]}. I'm just a little shorter than you, I'm only 5 feet tall. & 6.29  & 7.01  & 0     & 2/2   & 3/2   & 2/2 \\
    \midrule
    GPT-2 & \multicolumn{1}{l|}{I love florida.} & 4.00  & 8.32  & 0     & 3/3   & 2/2   & 1/1 \\
    \quad+PAL & I love florida. By the way, I have \textbf{a dog named pedro}. & 16.00 & 23.08 & 1     & 3/3   & 3/2   & 3/3 \\
    \midrule
    DialoGPT & Yes   & 0.00  & 0.00  & 0     & 1/1   & 1/1   & 1/1 \\
    \quad+PAL & I love florida. I have \textbf{a dog named pedro}. & 24.00 & 38.57 & 1     & 3/3   & 2/2   & 3/3 \\
    \midrule
    Llama 3.1 8B & I am in texas . I am a \textcolor{red}{nurse}  & 8.00  & 12.86 & -1    & 3/3   & 3/3   & 2/2 \\
    \quad+PAL & Very nice, I love florida. And I have \textbf{a dog named pedro}. Do you have any pet? & 20.00 & 19.17 & 1     & 3/3   & 3/3   & 3/3 \\
    \bottomrule
    \end{tabular}
    }

    \caption{\label{tab:case_study}Case Study Comparing Our Framework with Several Strong Baselines. \textbf{Bold} text indicates alignment with personas, while \textcolor{red}{red} text highlights less preferred responses that either contradict the personas or introduce unsupported details. PerCons refers to persona consistency. Human evaluation results are shown in the format A/B, where A and B correspond to the scores assigned by the first and second annotators, respectively.}
  \label{tab:engcase}
\end{table*}

\subsection{Impacts of Different Inference Strategies}
To demonstrate the effectiveness of our \textit{Select then Generate} inference strategy in filtering noisy and redundant personas, we compared it with two straightforward inference methods, \textbf{Random} and \textbf{Not Select}, on the Original PERSONA-CHAT dataset. In the Random strategy, the model-selected persona is replaced with a randomly chosen one. In contrast, the Not Select strategy bypasses the persona selection process entirely, using the full personas for dialogue generation. The results in Figure \ref{fig_generatethenselect} show that our strategy significantly outperforms these alternatives, underscoring the importance of selecting dialogue-informed personas for effective personalized dialogue generation.

\subsection{Impacts of Alignment Training Steps}
\begin{figure}[!t]
\begin{center}
\includegraphics[width=0.4\textwidth]{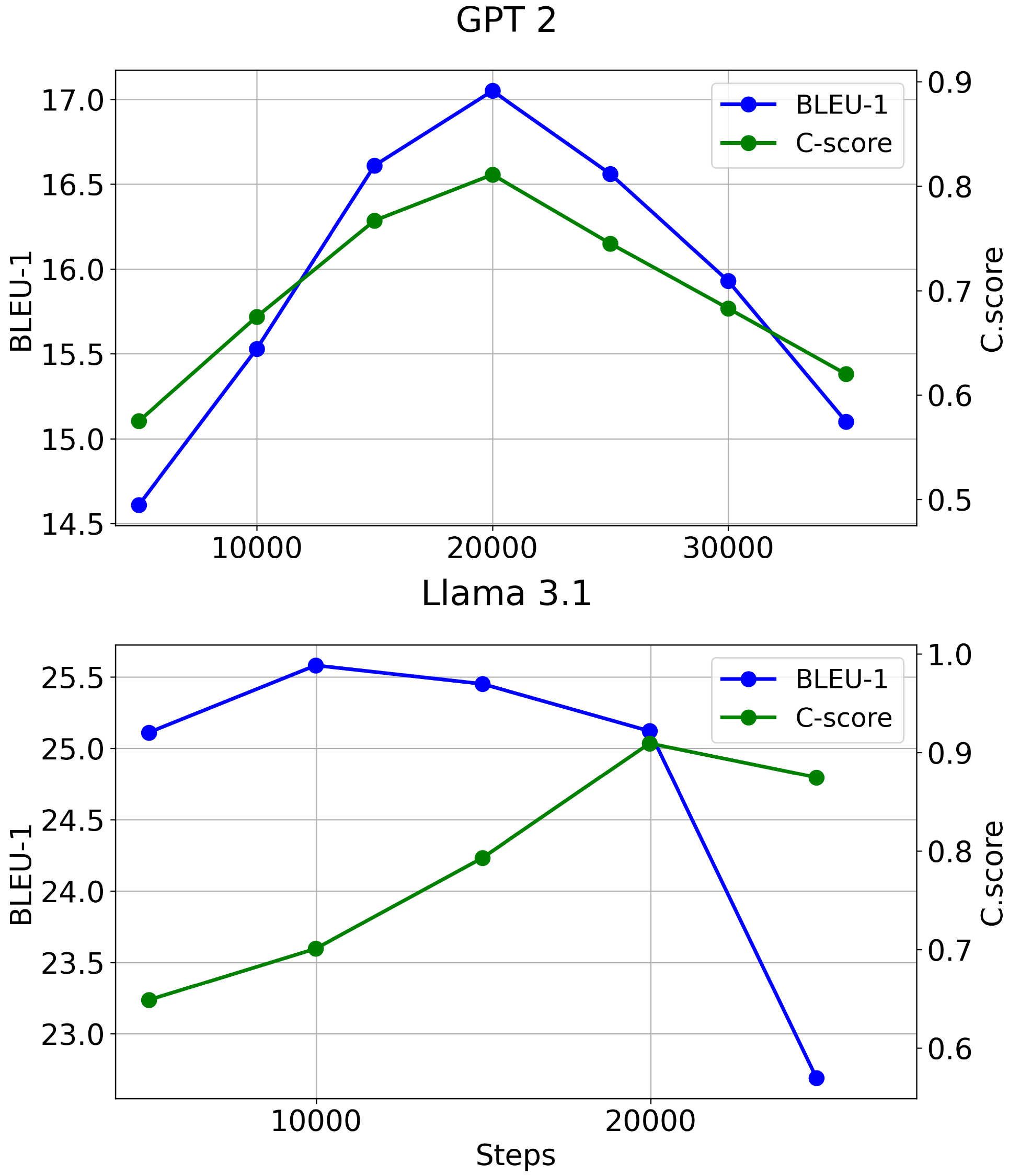} 
\caption{Influence of Persona Alignment Training Steps on the PERSONA-CHAT Dataset.}
\label{fig_iat}
\end{center}
\end{figure}
Previous research highlights the importance of appropriate alignment steps in training models. Insufficient alignment steps can prevent models from fully adapting to user preferences, while excessive steps might lead to overfitting and degrade performance. As demonstrated in Figure \ref{fig_iat}, our experimental results on both GPT 2 and Llama 3.1 8B confirm this, showing that proper calibration of alignment steps is crucial for model effectiveness. Our experiments emphasize the need to find the optimal number of alignment steps to ensure models align with personas without compromising their general language understanding abilities.

\subsection{Case Study}\label{casestudy_sec}
Table \ref{tab:case_study} presents an example of our framework alongside several baseline models. From the results, we observe that:

\begin{enumerate} 
\item The PAL framework generates more personalized responses compared to foundation models. Specifically, GPT-2 and Llama 3.1 8B models failed to produce responses that align with the given personas. In contrast, our framework consistently generates better-tailored responses to the specified personas. These findings demonstrate the effectiveness of our framework in personalizing responses according to the given personas.

\item As highlighted in red of Table \ref{tab:case_study}, both GPT-4o Mini and Llama 3.1 8B generate responses that include elements absent from the dialogue history and user personas, which are regarded as instances of hallucination. For example, the reference to a ``nurse'' in the Llama 3.1 8B response is an inference made by the model without any supporting evidence from the provided personas or dialogue context, which can be considered a hallucination. Moreover, Gemini Flash generates responses including placeholder ``[mention your specific area in healthcare]''. This indicates a significant lack of coherence and relevance, as the generated content does not reflect the user's preferences or the context of the dialogue.

\item Although large-scale models typically incorporate personas, their responses are often blunt and lack fluency. For example, phrases like ``film'' in GPT-3.5 or ``5 feet tall'' in Gemini Flash are completely unrelated to the dialogue. Our framework, which appropriately mentions relevant elements ``a dog named pedro'', is better suited to the dialogue context involving veterinarians and doctors.

\end{enumerate}

\section{Related Works}
\subsection{Personalized Dialogue Agents}

Personalized dialogue, which integrates personalized information and dialogue history, has been identified as essential for achieving human-like conversations and has attracts increasing research interest. Typical works use data-driven methods \cite{mazare2018training,song2021bob,chen0DK0W23,zhang2018personalizing,zheng2020pre,liuCFMM23}.
However, these methods usually overlook the intricate relationship between personas and dialogue context. To bridge this gap, some studies have aimed to capture the essential connection between personas and dialogue context \cite{PosokhovMMMM22, Huang0KL00T23,tangWFZHHH23}.

Concurrently, various studies \cite{shum2020sketch,song2021bob,liu2022improving,chen0DK0W23,tangWFZHHH23,Chen0Y023} focus on enhancing the consistency between responses and the corresponding personas. \citet{li2023learning} propose a coarse-to-fine persona-aware training framework to improve the persona consistency of a dialogue agent progressively. \citet{tangWFZHHH23} combines sparse and dense persona descriptions with dialogue history to design a Contrastive Latent Variable-based model (CLV) for generating personalized responses. \citet{huang-etal-2024-selective} utilizes a trainable dense retriever and adaptive soft prompts to enhance personalized dialogue generation in large language models by dynamically selecting relevant prompts based on conversational context. \citet{XuGWNW0W22} propose a long-term memory dialogue system based on an explicit memory read–write mechanism. Their framework consists of three components: a persona extractor, a long-term persona memory, and a generation module. This design demonstrates the effectiveness of selecting from multiple personas and serves as inspiration for our adoption of the ``Select-then-Generate'' strategy.

Although these methods have achieved significant success, they rely on token-level training objectives, such as Masked Prediction (MP) and Next Token Prediction (NTP). However, as shown in our experiments, these objectives fall short of fully capturing the alignment between personas and responses. In contrast, our framework directly optimizes this alignment, a crucial factor for effective personalized dialogue generation. Unlike training-based approaches, \citet{ZhouPSC23} focus on the inference phase, where they oversample a large number of candidate responses and then post-evaluate them to find a suitable match. However, without alignment-focused training, their method has a lower performance upper bound than ours, as demonstrated by our experimental results.

\subsection{Agent Alignment}
Aligning language models with human intentions has become a crucial focus within the field of large language models. To this end, a variety of methods have been developed. For example, techniques such as RLHF \cite{ouyang0JAWMZASR22}, RLAIF \cite{lee2023rlaif}, DPO \cite{rafailov2023direct}, and Decision Transformer \cite{chen2021decision} have been proposed to direct LLMs towards desired outcomes. In exploring the critical aspect of alignment, LIMA \cite{zhou2023lima} investigated the hypothesis that LLMs inherently develop knowledge and capabilities during their pre-training phase. Similarly, InstructionGPT-4 \cite{wei2023instructiongpt} demonstrated the potential of using a minimal yet high-quality dataset to achieve significant improvements. Setting our work apart, we specifically target the domain of personalized dialogue agents, addressing the unique challenges that arise in this context. Unlike the broader approaches to model alignment, we focus on enhancing the interaction between dialogue agents and users by ensuring the generated responses are not only contextually appropriate but also deeply personalized.

\section{Conclusion}
In this work, we propose a Persona-Aware Alignment Framework (PAL) featuring a two-stage training strategy and a \textit{Select then Generate} inference strategy. Unlike previous studies that rely on token-level objectives, our framework directly optimizes the alignment between generated responses and given personas, effectively addressing practical challenges such as quantifying alignment between responses and personas, bridging the gap between personalized dialogue data and pre-training data, and managing noisy persona information. Our experiments demonstrate that PAL significantly and consistently outperforms state-of-the-art personalized dialogue methods and closed-source commercial large language models. Notably, the framework’s consistent performance across different languages and multiple foundation models underscores its generalizability and broad applicability.

\normalem
\section*{Acknowledgments}
We would like to thank the anonymous reviewers for their insightful comments. This work is supported by the NSFC (No. 62206126, 62376120) and the Fundamental Research Funds for the Central Universities (No. 022114380016).

\bibliography{tacl2021}

\begin{thebibliography}{47}
\expandafter\ifx\csname natexlab\endcsname\relax\def\natexlab#1{#1}\fi

\bibitem[{Brown et~al.(2020)Brown, Mann, Ryder, Subbiah, Kaplan, Dhariwal, Neelakantan, Shyam, Sastry, Askell et~al.}]{brown2020language}
Tom Brown, Benjamin Mann, Nick Ryder, Melanie Subbiah, Jared~D Kaplan, Prafulla Dhariwal, Arvind Neelakantan, Pranav Shyam, Girish Sastry, Amanda Askell, et~al. 2020.
\newblock Language models are few-shot learners.
\newblock \emph{Advances in neural information processing systems}, 33:1877--1901.

\bibitem[{Chen et~al.(2023{\natexlab{a}})Chen, Wang, Deng, Kwan, Wang, and Wong}]{chen0DK0W23}
Liang Chen, Hongru Wang, Yang Deng, Wai{-}Chung Kwan, Zezhong Wang, and Kam{-}Fai Wong. 2023{\natexlab{a}}.
\newblock Towards robust personalized dialogue generation via order-insensitive representation regularization.
\newblock In \emph{Findings of the Association for Computational Linguistics: {ACL} 2023, Toronto, Canada, July 9-14, 2023}, pages 7337--7345. Association for Computational Linguistics.

\bibitem[{Chen et~al.(2021)Chen, Lu, Rajeswaran, Lee, Grover, Laskin, Abbeel, Srinivas, and Mordatch}]{chen2021decision}
Lili Chen, Kevin Lu, Aravind Rajeswaran, Kimin Lee, Aditya Grover, Misha Laskin, Pieter Abbeel, Aravind Srinivas, and Igor Mordatch. 2021.
\newblock Decision transformer: Reinforcement learning via sequence modeling.
\newblock \emph{Advances in neural information processing systems}, 34:15084--15097.

\bibitem[{Chen et~al.(2023{\natexlab{b}})Chen, Wang, Yu, and Zhang}]{Chen0Y023}
Ruijun Chen, Jin Wang, Liang{-}Chih Yu, and Xuejie Zhang. 2023{\natexlab{b}}.
\newblock \href {https://doi.org/10.1609/AAAI.V37I11.26489} {Learning to memorize entailment and discourse relations for persona-consistent dialogues}.
\newblock In \emph{Thirty-Seventh {AAAI} Conference on Artificial Intelligence, {AAAI} 2023, Thirty-Fifth Conference on Innovative Applications of Artificial Intelligence, {IAAI} 2023, Thirteenth Symposium on Educational Advances in Artificial Intelligence, {EAAI} 2023, Washington, DC, USA, February 7-14, 2023}, pages 12653--12661. {AAAI} Press.

\bibitem[{Dubey et~al.(2024)Dubey, Jauhri, Pandey, Kadian, Al-Dahle, Letman, Mathur, Schelten, Yang, Fan et~al.}]{dubey2024llama}
Abhimanyu Dubey, Abhinav Jauhri, Abhinav Pandey, Abhishek Kadian, Ahmad Al-Dahle, Aiesha Letman, Akhil Mathur, Alan Schelten, Amy Yang, Angela Fan, et~al. 2024.
\newblock The llama 3 herd of models.
\newblock \emph{arXiv preprint arXiv:2407.21783}.

\bibitem[{Glockner et~al.(2018)Glockner, Shwartz, and Goldberg}]{GlocknerSG18}
Max Glockner, Vered Shwartz, and Yoav Goldberg. 2018.
\newblock \href {https://doi.org/10.18653/V1/P18-2103} {Breaking {NLI} systems with sentences that require simple lexical inferences}.
\newblock In \emph{Proceedings of the 56th Annual Meeting of the Association for Computational Linguistics, {ACL} 2018, Melbourne, Australia, July 15-20, 2018, Volume 2: Short Papers}, pages 650--655. Association for Computational Linguistics.

\bibitem[{Han et~al.(2024)Han, Koh, Seo, Chang, and Sohn}]{han-etal-2024-psydial-personality}
Ji-Eun Han, Jun-Seok Koh, Hyeon-Tae Seo, Du-Seong Chang, and Kyung-Ah Sohn. 2024.
\newblock \href {https://aclanthology.org/2024.lrec-main.1166} {{PSYDIAL}: Personality-based synthetic dialogue generation using large language models}.
\newblock In \emph{Proceedings of the 2024 Joint International Conference on Computational Linguistics, Language Resources and Evaluation (LREC-COLING 2024)}, pages 13321--13331, Torino, Italia. ELRA and ICCL.

\bibitem[{Hu et~al.(2022)Hu, Shen, Wallis, Allen{-}Zhu, Li, Wang, Wang, and Chen}]{HuSWALWWC22}
Edward~J. Hu, Yelong Shen, Phillip Wallis, Zeyuan Allen{-}Zhu, Yuanzhi Li, Shean Wang, Lu~Wang, and Weizhu Chen. 2022.
\newblock \href {https://openreview.net/forum?id=nZeVKeeFYf9} {Lora: Low-rank adaptation of large language models}.
\newblock In \emph{The Tenth International Conference on Learning Representations, {ICLR} 2022, Virtual Event, April 25-29, 2022}. OpenReview.net.

\bibitem[{Hu et~al.(2020)Hu, Richardson, Xu, Li, Kuebler, and Moss}]{ocnli}
Hai Hu, Kyle Richardson, Liang Xu, Lu~Li, Sandra Kuebler, and Larry Moss. 2020.
\newblock \href {https://arxiv.org/abs/2010.05444} {Ocnli: Original chinese natural language inference}.
\newblock In \emph{Findings of EMNLP}.

\bibitem[{Huang et~al.(2024)Huang, Liu, Ko, Wu, Wang, Zhang, and Tang}]{huang-etal-2024-selective}
Qiushi Huang, Xubo Liu, Tom Ko, Bo~Wu, Wenwu Wang, Yu~Zhang, and Lilian Tang. 2024.
\newblock \href {https://doi.org/10.18653/v1/2024.findings-acl.959} {Selective prompting tuning for personalized conversations with {LLM}s}.
\newblock In \emph{Findings of the Association for Computational Linguistics: ACL 2024}, pages 16212--16226, Bangkok, Thailand. Association for Computational Linguistics.

\bibitem[{Huang et~al.(2023)Huang, Zhang, Ko, Liu, Wu, Wang, and Tang}]{Huang0KL00T23}
Qiushi Huang, Yu~Zhang, Tom Ko, Xubo Liu, Bo~Wu, Wenwu Wang, and H.~Lilian Tang. 2023.
\newblock Personalized dialogue generation with persona-adaptive attention.
\newblock In \emph{Thirty-Seventh {AAAI} Conference on Artificial Intelligence, {AAAI} 2023, Thirty-Fifth Conference on Innovative Applications of Artificial Intelligence, {IAAI} 2023, Thirteenth Symposium on Educational Advances in Artificial Intelligence, {EAAI} 2023, Washington, DC, USA, February 7-14, 2023}, pages 12916--12923. {AAAI} Press.

\bibitem[{Kenton and Toutanova(2019)}]{kenton2019bert}
Jacob Devlin Ming-Wei~Chang Kenton and Lee~Kristina Toutanova. 2019.
\newblock Bert: Pre-training of deep bidirectional transformers for language understanding.
\newblock In \emph{Proceedings of NAACL-HLT}, pages 4171--4186.

\bibitem[{Kim et~al.(2019)Kim, Kang, and Kwak}]{KimKK19}
Seonhoon Kim, Inho Kang, and Nojun Kwak. 2019.
\newblock \href {https://doi.org/10.1609/AAAI.V33I01.33016586} {Semantic sentence matching with densely-connected recurrent and co-attentive information}.
\newblock In \emph{The Thirty-Third {AAAI} Conference on Artificial Intelligence, {AAAI} 2019, The Thirty-First Innovative Applications of Artificial Intelligence Conference, {IAAI} 2019, The Ninth {AAAI} Symposium on Educational Advances in Artificial Intelligence, {EAAI} 2019, Honolulu, Hawaii, USA, January 27 - February 1, 2019}, pages 6586--6593. {AAAI} Press.

\bibitem[{Lee et~al.(2023)Lee, Phatale, Mansoor, Lu, Mesnard, Bishop, Carbune, and Rastogi}]{lee2023rlaif}
Harrison Lee, Samrat Phatale, Hassan Mansoor, Kellie Lu, Thomas Mesnard, Colton Bishop, Victor Carbune, and Abhinav Rastogi. 2023.
\newblock Rlaif: Scaling reinforcement learning from human feedback with ai feedback.
\newblock \emph{arXiv preprint arXiv:2309.00267}.

\bibitem[{Lester et~al.(2021)Lester, Al-Rfou, and Constant}]{lester2021power}
Brian Lester, Rami Al-Rfou, and Noah Constant. 2021.
\newblock The power of scale for parameter-efficient prompt tuning.
\newblock In \emph{Proceedings of the 2021 Conference on Empirical Methods in Natural Language Processing}, pages 3045--3059.

\bibitem[{Li et~al.(2023)Li, Hu, Sun, Xing, Guo, Xie, and Peng}]{li2023learning}
Yunpeng Li, Yue Hu, Yajing Sun, Luxi Xing, Ping Guo, Yuqiang Xie, and Wei Peng. 2023.
\newblock Learning to know myself: A coarse-to-fine persona-aware training framework for personalized dialogue generation.
\newblock In \emph{Proceedings of the AAAI Conference on Artificial Intelligence}, volume~37, pages 13157--13165.

\bibitem[{Lin(2004)}]{lin2004rouge}
Chin-Yew Lin. 2004.
\newblock Rouge: A package for automatic evaluation of summaries.
\newblock In \emph{Text summarization branches out}, pages 74--81.

\bibitem[{Liu et~al.(2023)Liu, Cho, Freedman, Ma, and May}]{liuCFMM23}
Shuai Liu, Hyundong Cho, Marjorie Freedman, Xuezhe Ma, and Jonathan May. 2023.
\newblock {RECAP:} retrieval-enhanced context-aware prefix encoder for personalized dialogue response generation.
\newblock In \emph{Proceedings of the 61st Annual Meeting of the Association for Computational Linguistics (Volume 1: Long Papers), {ACL} 2023, Toronto, Canada, July 9-14, 2023}, pages 8404--8419. Association for Computational Linguistics.

\bibitem[{Liu et~al.(2022)Liu, Wei, Liu, Mao, Fang, and Chen}]{liu2022improving}
Yifan Liu, Wei Wei, Jiayi Liu, Xianling Mao, Rui Fang, and Dangyang Chen. 2022.
\newblock Improving personality consistency in conversation by persona extending.
\newblock In \emph{Proceedings of the 31st ACM International Conference on Information \& Knowledge Management}, pages 1350--1359.

\bibitem[{Liu et~al.(2019)Liu, Ott, Goyal, Du, Joshi, Chen, Levy, Lewis, Zettlemoyer, and Stoyanov}]{liu2019roberta}
Yinhan Liu, Myle Ott, Naman Goyal, Jingfei Du, Mandar Joshi, Danqi Chen, Omer Levy, Mike Lewis, Luke Zettlemoyer, and Veselin Stoyanov. 2019.
\newblock Roberta: A robustly optimized bert pretraining approach.
\newblock \emph{arXiv preprint arXiv:1907.11692}.

\bibitem[{Madotto et~al.(2019)Madotto, Lin, Wu, and Fung}]{madotto2019personalizing}
Andrea Madotto, Zhaojiang Lin, Chien-Sheng Wu, and Pascale Fung. 2019.
\newblock Personalizing dialogue agents via meta-learning.
\newblock In \emph{Proceedings of the 57th annual meeting of the association for computational linguistics}, pages 5454--5459.

\bibitem[{Mazare et~al.(2018)Mazare, Humeau, Raison, and Bordes}]{mazare2018training}
Pierre-Emmanuel Mazare, Samuel Humeau, Martin Raison, and Antoine Bordes. 2018.
\newblock Training millions of personalized dialogue agents.
\newblock In \emph{Proceedings of the 2018 Conference on Empirical Methods in Natural Language Processing}, pages 2775--2779.

\bibitem[{McTear(2022)}]{mctear2022cai}
Michael McTear. 2022.
\newblock \emph{Conversational ai: Dialogue systems, conversational agents, and chatbots}.
\newblock Springer Nature.

\bibitem[{Ouyang et~al.(2022)Ouyang, Wu, Jiang, Almeida, Wainwright, Mishkin, Zhang, Agarwal, Slama, Ray, Schulman, Hilton, Kelton, Miller, Simens, Askell, Welinder, Christiano, Leike, and Lowe}]{ouyang0JAWMZASR22}
Long Ouyang, Jeffrey Wu, Xu~Jiang, Diogo Almeida, Carroll~L. Wainwright, Pamela Mishkin, Chong Zhang, Sandhini Agarwal, Katarina Slama, Alex Ray, John Schulman, Jacob Hilton, Fraser Kelton, Luke Miller, Maddie Simens, Amanda Askell, Peter Welinder, Paul~F. Christiano, Jan Leike, and Ryan Lowe. 2022.
\newblock Training language models to follow instructions with human feedback.
\newblock In \emph{NeurIPS}.

\bibitem[{Papineni et~al.(2002)Papineni, Roukos, Ward, and Zhu}]{papineni2002bleu}
Kishore Papineni, Salim Roukos, Todd Ward, and Wei-Jing Zhu. 2002.
\newblock Bleu: a method for automatic evaluation of machine translation.
\newblock In \emph{Proceedings of the 40th annual meeting of the Association for Computational Linguistics}, pages 311--318.

\bibitem[{Pei et~al.(2021)Pei, Ren, and de~Rijke}]{pei2021cooperative}
Jiahuan Pei, Pengjie Ren, and Maarten de~Rijke. 2021.
\newblock A cooperative memory network for personalized task-oriented dialogue systems with incomplete user profiles.
\newblock In \emph{Proceedings of the Web Conference 2021}, pages 1552--1561.

\bibitem[{Posokhov et~al.(2022)Posokhov, Matveeva, Makhnytkina, Matveev, and Matveev}]{PosokhovMMMM22}
Pavel Posokhov, Anastasia Matveeva, Olesia Makhnytkina, Anton Matveev, and Yuri Matveev. 2022.
\newblock Personalizing retrieval-based dialogue agents.
\newblock In \emph{Speech and Computer - 24th International Conference, {SPECOM} 2022, Gurugram, India, November 14-16, 2022, Proceedings}, volume 13721 of \emph{Lecture Notes in Computer Science}, pages 554--566. Springer.

\bibitem[{Radford et~al.(2019)Radford, Wu, Child, Luan, Amodei, Sutskever et~al.}]{radford2019language}
Alec Radford, Jeffrey Wu, Rewon Child, David Luan, Dario Amodei, Ilya Sutskever, et~al. 2019.
\newblock Language models are unsupervised multitask learners.
\newblock \emph{OpenAI blog}, 1(8):9.

\bibitem[{Rafailov et~al.(2023)Rafailov, Sharma, Mitchell, Ermon, Manning, and Finn}]{rafailov2023direct}
Rafael Rafailov, Archit Sharma, Eric Mitchell, Stefano Ermon, Christopher~D Manning, and Chelsea Finn. 2023.
\newblock Direct preference optimization: Your language model is secretly a reward model.
\newblock \emph{arXiv preprint arXiv:2305.18290}.

\bibitem[{Sarikaya et~al.(2016)Sarikaya, Crook, Marin, Jeong, Robichaud, Celikyilmaz, Kim, Rochette, Khan, Liu et~al.}]{sarikaya2016overview}
Ruhi Sarikaya, Paul~A Crook, Alex Marin, Minwoo Jeong, Jean-Philippe Robichaud, Asli Celikyilmaz, Young-Bum Kim, Alexandre Rochette, Omar~Zia Khan, Xiaohu Liu, et~al. 2016.
\newblock An overview of end-to-end language understanding and dialog management for personal digital assistants.
\newblock In \emph{2016 ieee spoken language technology workshop (slt)}, pages 391--397. IEEE.

\bibitem[{Schulman et~al.(2017)Schulman, Wolski, Dhariwal, Radford, and Klimov}]{schulman2017proximal}
John Schulman, Filip Wolski, Prafulla Dhariwal, Alec Radford, and Oleg Klimov. 2017.
\newblock Proximal policy optimization algorithms.
\newblock \emph{arXiv preprint arXiv:1707.06347}.

\bibitem[{Shum et~al.(2020)Shum, Zheng, Kry{\'s}ci{\'n}ski, Xiong, and Socher}]{shum2020sketch}
Michael Shum, Stephan Zheng, Wojciech Kry{\'s}ci{\'n}ski, Caiming Xiong, and Richard Socher. 2020.
\newblock Sketch-fill-ar: A persona-grounded chit-chat generation framework.
\newblock In \emph{Proceedings of the 2nd Workshop on Natural Language Processing for Conversational AI}, pages 118--131.

\bibitem[{Singh and Beniwal(2022)}]{singh2022survey}
Satwinder Singh and Himanshu Beniwal. 2022.
\newblock A survey on near-human conversational agents.
\newblock \emph{Journal of King Saud University-Computer and Information Sciences}, 34(10):8852--8866.

\bibitem[{Song et~al.(2021)Song, Wang, Zhang, Zhang, and Liu}]{song2021bob}
Haoyu Song, Yan Wang, Kaiyan Zhang, Weinan Zhang, and Ting Liu. 2021.
\newblock Bob: Bert over bert for training persona-based dialogue models from limited personalized data.
\newblock In \emph{Proceedings of the 59th Annual Meeting of the Association for Computational Linguistics and the 11th International Joint Conference on Natural Language Processing (Volume 1: Long Papers)}, pages 167--177.

\bibitem[{Tang et~al.(2023)Tang, Wang, Fang, Zhao, Huang, He, and Hou}]{tangWFZHHH23}
Yihong Tang, Bo~Wang, Miao Fang, Dongming Zhao, Kun Huang, Ruifang He, and Yuexian Hou. 2023.
\newblock Enhancing personalized dialogue generation with contrastive latent variables: Combining sparse and dense persona.
\newblock In \emph{Proceedings of the 61st Annual Meeting of the Association for Computational Linguistics (Volume 1: Long Papers), {ACL} 2023, Toronto, Canada, July 9-14, 2023}, pages 5456--5468. Association for Computational Linguistics.

\bibitem[{Wei et~al.(2023)Wei, Jiang, Huang, and Sun}]{wei2023instructiongpt}
Lai Wei, Zihao Jiang, Weiran Huang, and Lichao Sun. 2023.
\newblock Instructiongpt-4: A 200-instruction paradigm for fine-tuning minigpt-4.
\newblock \emph{arXiv preprint arXiv:2308.12067}.

\bibitem[{Wolf et~al.(2020)Wolf, Debut, Sanh, Chaumond, Delangue, Moi, Cistac, Rault, Louf, Funtowicz et~al.}]{wolf2020transformers}
Thomas Wolf, Lysandre Debut, Victor Sanh, Julien Chaumond, Clement Delangue, Anthony Moi, Pierric Cistac, Tim Rault, R{\'e}mi Louf, Morgan Funtowicz, et~al. 2020.
\newblock Transformers: State-of-the-art natural language processing.
\newblock In \emph{Proceedings of the 2020 conference on empirical methods in natural language processing: system demonstrations}, pages 38--45.

\bibitem[{Wu et~al.(2021)Wu, Zheng, Mao, and Huang}]{wu2021transferable}
Chen~Henry Wu, Yinhe Zheng, Xiaoxi Mao, and Minlie Huang. 2021.
\newblock Transferable persona-grounded dialogues via grounded minimal edits.
\newblock In \emph{Proceedings of the 2021 Conference on Empirical Methods in Natural Language Processing}, pages 2368--2382.

\bibitem[{Xiong et~al.(2024)Xiong, Dong, Ye, Wang, Zhong, Ji, Jiang, and Zhang}]{0015DYW0J0024}
Wei Xiong, Hanze Dong, Chenlu Ye, Ziqi Wang, Han Zhong, Heng Ji, Nan Jiang, and Tong Zhang. 2024.
\newblock \href {https://openreview.net/forum?id=c1AKcA6ry1} {Iterative preference learning from human feedback: Bridging theory and practice for {RLHF} under kl-constraint}.
\newblock In \emph{Forty-first International Conference on Machine Learning, {ICML} 2024, Vienna, Austria, July 21-27, 2024}. OpenReview.net.

\bibitem[{Xu et~al.(2020)Xu, Hu, Zhang, Li, Cao, Li, Xu, Sun, Yu, Yu, Tian, Dong, Liu, Shi, Cui, Li, Zeng, Wang, Xie, Li, Patterson, Tian, Zhang, Zhou, Liu, Zhao, Zhao, Yue, Zhang, Yang, Richardson, and Lan}]{XuHZLCLXSYYTDLS20}
Liang Xu, Hai Hu, Xuanwei Zhang, Lu~Li, Chenjie Cao, Yudong Li, Yechen Xu, Kai Sun, Dian Yu, Cong Yu, Yin Tian, Qianqian Dong, Weitang Liu, Bo~Shi, Yiming Cui, Junyi Li, Jun Zeng, Rongzhao Wang, Weijian Xie, Yanting Li, Yina Patterson, Zuoyu Tian, Yiwen Zhang, He~Zhou, Shaoweihua Liu, Zhe Zhao, Qipeng Zhao, Cong Yue, Xinrui Zhang, Zhengliang Yang, Kyle Richardson, and Zhenzhong Lan. 2020.
\newblock \href {https://doi.org/10.18653/V1/2020.COLING-MAIN.419} {{CLUE:} {A} chinese language understanding evaluation benchmark}.
\newblock In \emph{Proceedings of the 28th International Conference on Computational Linguistics, {COLING} 2020, Barcelona, Spain (Online), December 8-13, 2020}, pages 4762--4772. International Committee on Computational Linguistics.

\bibitem[{Xu et~al.(2022)Xu, Gou, Wu, Niu, Wu, Wang, and Wang}]{XuGWNW0W22}
Xinchao Xu, Zhibin Gou, Wenquan Wu, Zheng{-}Yu Niu, Hua Wu, Haifeng Wang, and Shihang Wang. 2022.
\newblock \href {https://doi.org/10.18653/V1/2022.FINDINGS-ACL.207} {Long time no see! open-domain conversation with long-term persona memory}.
\newblock In \emph{Findings of the Association for Computational Linguistics: {ACL} 2022, Dublin, Ireland, May 22-27, 2022}, pages 2639--2650. Association for Computational Linguistics.

\bibitem[{Zhang et~al.(2018{\natexlab{a}})Zhang, Dinan, Urbanek, Szlam, Kiela, and Weston}]{zhang2018personalizing}
Saizheng Zhang, Emily Dinan, Jack Urbanek, Arthur Szlam, Douwe Kiela, and Jason Weston. 2018{\natexlab{a}}.
\newblock Personalizing dialogue agents: I have a dog, do you have pets too?
\newblock In \emph{Proceedings of the 56th Annual Meeting of the Association for Computational Linguistics (Volume 1: Long Papers)}, pages 2204--2213.

\bibitem[{Zhang et~al.(2018{\natexlab{b}})Zhang, Galley, Gao, Gan, Li, Brockett, and Dolan}]{zhang2018generating}
Yizhe Zhang, Michel Galley, Jianfeng Gao, Zhe Gan, Xiujun Li, Chris Brockett, and Bill Dolan. 2018{\natexlab{b}}.
\newblock Generating informative and diverse conversational responses via adversarial information maximization.
\newblock \emph{Advances in Neural Information Processing Systems}, 31.

\bibitem[{Zhang et~al.(2020)Zhang, Sun, Galley, Chen, Brockett, Gao, Gao, Liu, and Dolan}]{zhang2020dialogpt}
Yizhe Zhang, Siqi Sun, Michel Galley, Yen-Chun Chen, Chris Brockett, Xiang Gao, Jianfeng Gao, Jingjing Liu, and William~B Dolan. 2020.
\newblock Dialogpt: Large-scale generative pre-training for conversational response generation.
\newblock In \emph{Proceedings of the 58th Annual Meeting of the Association for Computational Linguistics: System Demonstrations}, pages 270--278.

\bibitem[{Zheng et~al.(2020)Zheng, Zhang, Huang, and Mao}]{zheng2020pre}
Yinhe Zheng, Rongsheng Zhang, Minlie Huang, and Xiaoxi Mao. 2020.
\newblock A pre-training based personalized dialogue generation model with persona-sparse data.
\newblock In \emph{Proceedings of the AAAI Conference on Artificial Intelligence}, volume~34, pages 9693--9700.

\bibitem[{Zhou et~al.(2023{\natexlab{a}})Zhou, Liu, Xu, Iyer, Sun, Mao, Ma, Efrat, Yu, Yu et~al.}]{zhou2023lima}
Chunting Zhou, Pengfei Liu, Puxin Xu, Srini Iyer, Jiao Sun, Yuning Mao, Xuezhe Ma, Avia Efrat, Ping Yu, Lili Yu, et~al. 2023{\natexlab{a}}.
\newblock Lima: Less is more for alignment.
\newblock \emph{arXiv preprint arXiv:2305.11206}.

\bibitem[{Zhou et~al.(2023{\natexlab{b}})Zhou, Pang, Shen, and Cheng}]{ZhouPSC23}
Junkai Zhou, Liang Pang, Huawei Shen, and Xueqi Cheng. 2023{\natexlab{b}}.
\newblock \href {https://doi.org/10.18653/V1/2023.ACL-LONG.553} {Simoap: Improve coherence and consistency in persona-based dialogue generation via over-sampling and post-evaluation}.
\newblock In \emph{Proceedings of the 61st Annual Meeting of the Association for Computational Linguistics (Volume 1: Long Papers), {ACL} 2023, Toronto, Canada, July 9-14, 2023}, pages 9945--9959. Association for Computational Linguistics.

\end{thebibliography}
\bibliographystyle{acl_natbib}

\iftaclpubformat

\onecolumn

\appendix
\section{Details about Annotations}\label{sec:annota}
The human evaluation is conducted by two postgraduate researchers specializing in natural language processing. Both annotators are native Chinese speakers and are proficient in English reading and writing, having passed the College English Test. Moreover, the annotators were provided with 50 randomly selected responses annotated by native English and Chinese speakers as reference points for quality and consistency. The inter-annotator agreement for the gold data, measured by the Fleiss Kappa coefficient, is 0.62, indicating moderate agreement among annotators. The annotations for PERSONA-CHAT dataset are performed using the original PERSONA-CHAT version.

\section{Human Evaluation Results on Baidu-Chat Dataset}
\label{sec:baiduchathuman}
As shown in Table \ref{tab:baiduhuman}, the human evaluation results on the Baidu-Chat dataset are consistent with the observations mentioned in the Overall Results section (Section \ref{overallresults}).

\begin{table}[!ht]
  \centering
\begin{tabular}{l|ccc}
\toprule
\multicolumn{1}{c|}{Models} & Fluency & Coherence & Persona Consistency \\
\midrule
ORIG  & \textbf{2.76}  & \textbf{2.46}  & \textbf{2.46} \\
CLV   & \underline{2.44}  & 1.97  & 1.51 \\
LMEDR & 2.30  & \underline{2.14}  & \underline{2.17} \\
\midrule
GPT-3.5 & 2.47  & 2.59  & 2.59 \\
\quad+Few-Shot & 2.68  & 2.61  & \underline{2.69} \\
GPT 4o mini & \underline{2.77}  & \textbf{2.75} & 2.60 \\
\quad+Few-Shot & 2.51  & 2.59  & \textbf{2.70} \\
Gemini Flash & 2.56  & \underline{2.72}  & \underline{2.69} \\
\quad+Few-Shot & \textbf{2.81} & 2.66  & 2.65 \\
\midrule
GPT-2 & 1.45  & 1.34  & 1.39 \\
\quad +FineTuning & 2.21  & 2.14  & 1.99 \\
\quad +Prompting & \underline{2.26} & 2.11  & \underline{2.39} \\
\quad+SimOAP & 2.08  & 1.65  & 1.79 \\
 \quad+SPT & \textbf{2.34} & \underline{2.29} & 2.19 \\
\quad+PAL(ours) & \textbf{2.34} & \textbf{2.47*} & \textbf{2.57*} \\
\midrule
Llama 3.1 8B & 2.83  & 2.41  & 2.55 \\
\quad+Few-Shot & 2.57  & 1.76  & 2.43 \\
\quad+Prompting & 2.85  & \underline{2.79} & 2.68 \\
\quad+SimOAP & 2.74  & 2.14  & \underline{2.69} \\
 \quad+SPT & \underline{2.88} & 2.48  & 2.63 \\
\quad+PAL(ours) & \textbf{2.89} & \textbf{2.80} & \textbf{2.74*} \\
\bottomrule
\end{tabular}%
\caption{\label{tab:baiduhuman}Human evaluation results on Baidu-Chat Dataset. The best results are in \textbf{bold} while the secondary results are marked \underline{underlined}. * denotes statistically significant (p<0.05) improvements over baselines on the same foundation models.}
\end{table}%

\section{Prompts for Closed-source LLMs}\label{prompt_closellm}
For our experiments with closed-source large language models on personalized dialogue tasks, we use the following standardized prompt:
\begin{tcolorbox}
\small
    \textbf{Prompt for closed-source large language models}\\
    The user's persona is described with: \textless personas\textgreater. \\Please generate a response to the dialogue. \\Dialogue context: \textless dialogue context\textgreater. \\Response: \textless response\textgreater
\end{tcolorbox}

This prompt is exactly the same as the prompt for Persona-Enhanced Dialogue Generation in Section \ref{prompt_construction} and is consistently applied across all closed-source model evaluations to ensure comparability of results.

\section{Examples from Baidu-Chat}
In this section, we present responses from different models on the Baidu-Chat dataset, along with their corresponding automatic metric scores and human evaluation results. These results are provided in Table \ref{tab:baiducase}. The selected case from the Baidu-Chat dataset is consistent with the findings discussed in the Case Study section (Section 5.4). Furthermore, the response quality in this case is higher than that observed in the PERSONA-CHAT dataset, which aligns with the superior performance reflected in both the automatic metrics and human evaluation results.

\begin{CJK}{UTF8}{gbsn}
\begin{table}[htbp]
  \centering
  \resizebox{\textwidth}{!}{
    \begin{tabular}{l|p{24.665em}|ccc|ccc}
    \toprule
    \multicolumn{8}{c}{Personas} \\
    \midrule
    \multicolumn{8}{p{58.665em}}{我是最早试用微博的人。(I was one of the earliest users of Weibo.)\newline{}我是个配音演员。(I am a voice actor.)\newline{}我曾经在漠河钓过鱼。(I once went fishing in Mohe.)} \\
    \midrule
    \multicolumn{8}{c}{Dialogue Context} \\
    \midrule
    \multicolumn{8}{p{58.665em}}{Person 1: 你现在不忙吗？(Are you not busy now?)\newline{}Person 2: 不忙的，我今天歇班儿。(No, I'm off work today.)\newline{}Person 1: 听你的意思，你已经工作了？(From what you're saying, you're already working?)\newline{}Person 2: 是的，我已经参加了工作，我是个配音演员。(Yes, I've already started working. I'm a voice actor.)\newline{}Person 1: 哇塞，你的工作听起来是真棒，那你的声音一定是很好听吧。(Wow, your job sounds amazing. Your voice must be really nice, right?)\newline{}Person 2: 我的声音还好吧，不过我们这个工作必须要保护好自己的嗓子。(My voice is okay, but in this line of work, we have to take good care of our throats.)\newline{}Person 1: 哦，好吧。感觉现在做什么工作都不容易呀。(Oh, I see. It seems like no job is easy these days.)} \\
    \midrule
    \midrule
    \multicolumn{1}{c|}{\multirow{2}[2]{*}{Models}} & \multirow{2}[2]{*}{Generated Responses} & \multicolumn{3}{c|}{Metrics} & \multicolumn{3}{c}{Human Evaluation Results} \\
          & \multicolumn{1}{c|}{} & BLEU-1 & ROUGE-L & C.score & Fluency & Coherence & PerCons \\
    \midrule
    GPT 3.5 & 是啊，确实不容易。(Yes, it's really not easy.) & 28.49 & 14.72 & -1    & 3/3   & 2/3   & 1/1 \\
    \midrule
    GPT 4o mini & 是啊，保护嗓子确实很重要，尤其是我们这种需要用声音来表达的职业。不过我觉得每个工作都有它的挑战和乐趣。对了，你有没有什么特别的兴趣爱好呢？(Yes, protecting your voice is really important, especially for professions like ours that rely on vocal expression. But I think every job has its own challenges and joys. By the way, do you have any special hobbies?) & 4.59  & 12.54 & 1     & 3/3   & 2/3   & 2/2 \\
    \midrule
    Gemini Flash & 确实不容易，不过我觉得你的经历很丰富，你曾经在\textbf{漠河钓过鱼}？那一定很有意思吧！分享一下你的经历呗，我很想听听。(It's really not easy, but I think your experiences are quite rich—you once went \textbf{fishing in Mohe}? That must have been really interesting! Share your story, I'd love to hear it.) & 6.74  & 14.41 & 1     & 3/3   & 3/3   & 3/3 \\
    \midrule
    GPT-2 & \multicolumn{1}{l|}{我们可以交个朋友吗？(Can we be friends?)} & 7.69  & 11.28 & 0     & 3/3   & 1/1   & 1/1 \\
    \quad+PAL & 是呀，\textbf{配音演员}工作不容易。(Yeah, being a \textbf{voice actor} isn’t easy.) & 15.38 & 9.95  & 1     & 2/2   & 3/2   & 3/3 \\
    \midrule
    Llama 3.1 8B & 是的，工作不容易，但是我觉得生活也不容易。(Yes, work isn’t easy, but I think life isn’t easy either.) & 19.79 & 14.10 & 1     & 3/3   & 3/2   & 1/1 \\
    \quad+PAL & 是啊，工作确实不容易。说起来，你有没有什么特别的经历呢？(Yes, work really isn’t easy. Speaking of which, do you have any special experiences?) & 29.09 & 24.18 & 1     & 3/3   & 3/3   & 3/2 \\
    \bottomrule
    \end{tabular}
    }
    \caption{Case Study Comparing Our Framework with Several Strong Baselines in the Baidu-Chat dataset. \textbf{Bold} text indicates alignment with personas. PerCons refers to persona consistency. Human evaluation results are shown in the format A/B, where A and B correspond to the scores assigned by the first and second annotators, respectively.}
  \label{tab:baiducase}
\end{table}
\end{CJK}

\end{document}